\documentclass[journal]{IEEEtranTIE}
\usepackage{graphicx}
\usepackage{cite}
\usepackage{picinpar}
\usepackage{amsmath}
\usepackage{amssymb}
\usepackage{url}
\usepackage{flushend}
\usepackage[utf8]{inputenc}
\usepackage{colortbl}
\usepackage{soul}
\usepackage{multirow}
\usepackage{pifont}
\usepackage{color}
\usepackage{alltt}
\usepackage[hidelinks]{hyperref}
\usepackage{enumerate}
\usepackage{siunitx}
\usepackage{breakurl}
\usepackage{epstopdf}
\usepackage{pbox}
\usepackage{algorithm}
\usepackage{algpseudocode}
\usepackage{booktabs} 
\usepackage{booktabs, multirow, array}
\usepackage{array}
\usepackage[compact]{titlesec}

\algtext*{EndIf}
\algtext*{EndWhile}

\titlespacing*{\section}{2pt}{*0.5}{*0.5}

\begin{document}
\bstctlcite{BSTcontrol_et_al}

\title{Physics-Embedded Neural ODEs for Sim-to-Real Edge Digital Twins of \\Hybrid Power Electronics Systems}

\vspace{-15pt}

\author{
	\vskip 1em
	
	Jialin Zheng, \emph{Member,~IEEE},
	Haoyu Wang, \emph {Graduate Student Member,~IEEE},
        Yangbin Zeng, \emph{Member,~IEEE},
        Di Mou, \emph{Member,~IEEE},
        Xin Zhang, \emph{Senior Member,~IEEE},
        Hong Li, \emph{Senior Member,~IEEE},
        Sergio Vazquez, \emph{Fellow,~IEEE},
        Leopoldo G. Franquelo, \emph{Life Fellow,~IEEE}
        \vspace{-15pt}

	\thanks{
	      Jialin Zheng and Haoyu Wang contributed equally to this work. 
       
		This paper was supported by the Guangdong Basic and Applied Basic Research Foundation under Grant 2025A1515011715, the National Natural Science Foundation of China under Grant 52237008, and the State Key Laboratory of Power System Operation and Control under Grant SKLD24KM27. \emph{(Corresponding author: Yangbin Zeng).}	
	}
}
\maketitle

\begin{abstract}

Edge Digital Twins (EDTs) are crucial for monitoring and control of Power Electronics Systems (PES). However, existing modeling approaches struggle to consistently capture continuously evolving hybrid dynamics that are inherent in PES, degrading Sim-to-Real generalization on resource-constrained edge devices. To address these challenges, this paper proposes a Physics-Embedded Neural ODEs (PENODE) that (i) embeds the hybrid operating mechanism as an event automaton to explicitly govern discrete switching and (ii) injects known governing ODE components directly into the neural parameterization of unmodeled dynamics. This unified design yields a differentiable end-to-end trainable architecture that preserves physical interpretability while reducing redundancy, and it supports a cloud-to-edge toolchain for efficient FPGA deployment. Experimental results demonstrate that PENODE achieves significantly higher accuracy in benchmarks in white-box, gray-box, and black-box scenarios, with a 75\% reduction in neuron count, validating that the proposed PENODE maintains physical interpretability, efficient edge deployment, and real-time control enhancement.

\end{abstract}


\markboth{IEEE TRANSACTIONS ON INDUSTRIAL ELECTRONICS}%
{}

\definecolor{limegreen}{rgb}{0.2, 0.8, 0.2}
\definecolor{forestgreen}{rgb}{0.13, 0.55, 0.13}
\definecolor{greenhtml}{rgb}{0.0, 0.5, 0.0}

\vspace{-15pt}
\section{Introduction}
 
\IEEEPARstart{P}{ower} electronics systems (PES) are the fundamental to drive efficient energy conversion \cite{10070105} but require precise and real-time monitoring and predictive analysis due to the ultra-high standards for reliability and performance \cite{9535421}. Digital Twin (DT), a high-fidelity virtual counterpart of a physical asset, presents a promising solution \cite{9784425}. However, it is difficult to implement cloud-or-server-based DTs with high communication latency and limited bandwidth in PES because the PES dynamics differ significantly from power grids \cite{9950705}.

With the emergence of edge computing, Edge DT (EDT) becomes a new paradigm to address the latency problem \cite{9921407}. By deploying a twin model on an edge node close to the PES, EDT enables real-time interaction with ultra-low latency, which not only satisfies the real-time monitoring demands but also allows the DT to evolve from a passive monitor into an active controller \cite{11077776}. However, it introduces a more formidable challenge that the model must possess robust transferability from offline simulators to real systems (Sim-to-Real). Any distortion between the model and the real system could lead to failures and pose severe safety risks \cite{9525187}.

In the literature, mainstream modeling methods, including the physics-based and data-driven methods, have fundamental limitations. On the one hand, physics-based models (e.g., state-space models) fail in real-world applications despite their accuracy in simulation environments \cite{plecsmanual}. They cannot automatically account for unmodeled dynamics or parameter drift, creating a gap between theory and reality \cite{4303291}. On the other hand, data-driven methods (typically black-box models) ignore physics priors and must learn from scratch, which demands immense training costs and vast datasets \cite{10057400, 7222462}. For example, \cite{7222462} required nearly one million data points to accurately model a dual active bridge (DAB) converter. Such heavy reliance on data and the disregard for physical laws make the Sim-to-Real transfer unreliable and inefficient. Consequently, neither physics-based nor data-driven approaches can independently solve the Sim-to-Real challenge. 

A promising path is to integrate physical principles with data-driven insights. However, such integration on resource-constrained edge devices faces challenges at three levels: architecture, methodology, and hardware implementation.

At the architectural level, the challenge originates from the nature that high-frequency switching of power semiconductors in PES creates multiple operating modes, each governed by its own continuous-time Ordinary Differential Equations (ODEs), and discrete topological jumps at switching instants \cite{10342664, 9819434}. To capture these hybrid dynamics, specialized event-driven frameworks emerged to explicitly handle state quantization and event scheduling \cite{10236507, 9832797}. However, these physics-centric approaches rely on switching assumptions and complete circuit model prior knowledge that is not easily transferable to data-driven models. Conversely, existing data-driven methods are typically designed for single-modal continuous systems \cite{DBLP}, which often evade direct modeling of the multi-modal dynamics and focus on sub-problems where continuous dynamics can be assumed \cite{7129353,10057400}. For example, \cite{7129353} uses a neuro-fuzzy inference system to relate the input voltage to the aging state of a DC filter capacitor and a fully connected neural network (NN) is used to map circuit parameters to system characteristics in \cite{10057400}. Thus, they can only implicitly learn hybrid dynamics using complex networks and large datasets \cite{9067098,9813409}. A NARX network in \cite{9067098} requires 4000 curated data entries to cover the modal transition in a Buck converter. Similarly, \cite{9813409} uses Hammerstein-Wiener structures to identify three-phase inverters but can only approximate different modes by increasing the model order and dataset size. To conclude, there lacks a top-level framework to efficiently describe the event-state hybrid nature.

At the methodological level, the data-driven methods used to model each mode are inherently discrete-time although Recurrent NNs (RNNs) and their variants (LSTMs, GRUs) are choices for time-series modeling \cite{9067098, 9255375, 10463542}. The step-by-step internal mechanism learns a mapping from one discrete state to the next, which faces a fundamental mismatch with PES governed by multiple ODEs with varying durations. Thus, a "discretization step dilemma" occurs, where no single fixed integration step is adequate because a large step size will miss short-lived modes and cause state errors while a small step size generates massive redundant computations during long-duration modes \cite{9832797}. Besides, discrete networks are numerical approximators of the state evolution trajectory at a fixed step size, instead of direct representations of the underlying physical laws (i.e., ODEs), which leads to the fact that a model trained at one step size cannot adapt to another step size \cite{NODE}. Although methods like Physics-Informed NN (PINN) and Physics-in-RNN (PRNN) attempt to incorporate physical knowledge, they fail to overcome the issue \cite{10463542, 9779551}. PINN apply physical laws only as a soft constraint with an external loss term and ignore specific topologies and operating rules of power electronics \cite{10463542}. PRNN combine a discretized known ODE model with a discrete NN, whose disparate natures prevent joint training and require a careful balance of conflicts in a two-stage process \cite{ 9779551}. Therefore, there are strong needs for an efficient approach that can achieve the deep integration and high computational efficiency. 

At the hardware implementation level, it is a big challenge to deploy data-driven models onto edge devices. Mainstream models are typically developed and evaluated in a cloud or high-performance computing environments with comprehensive software stacks (e.g., Python and TensorFlow) and powerful computational resources (CPUs and GPUs) \cite{10463542, 9779551,9067098, 9255375}. In contrast, edge devices (e.g., FPGA and Zynq SoC) are severely constrained in both memory and computational capabilities \cite{ 9832797}. 
Although tools like ONNX, FINN, and HLS4ML have been proposed to aid this transition to the edge, it remains exceedingly difficult to meet the sub-microsecond real-time requirements, especially for complex NN architectures \cite{9693757}. Furthermore, the data acquisition workflow is another critical bottleneck that the reliance on mere simulation data for training broadens the Sim-to-Real gap \cite{10463542, 9067098,9255375, 7222462}. However, collecting high-quality data from a physical converter under all operating conditions is prohibitively expensive, inefficient, and highly prone to error \cite{9779551,7129353}. Therefore, an efficient end-to-end Sim-to-Real deployment pipeline is strongly needed.

To address above challenges, this paper proposes a Physics-Embedded Neural ODEs (PENODE), which is designed to provide an end-to-end solution from theory to practice. Below are the key contributions of the method:

\begin{enumerate}
    \item An event-automata learning framework for multi-modal hybrid systems that explicitly identifies operating modes and models multi-modal dynamics to improve learning efficiency and accuracy for complex hybrid behaviors.

    \item A physics-embedded continuous-time neural-network model that injects known ODE primitives and learns only residual dynamics, yielding interpretable and redundancy-minimal Sim-to-Real transfer.
    
    \item A cloud-to-edge deployment workflow that facilitate the efficient transfer of the PENODE model to hardware through model optimization and hardware co-design, and satisfy the stringent real-time inference requirements.
\end{enumerate}

The remainder of this paper is organized as follows. Section \ref{SectionII} proposes the event-automata framework. In Section \ref{SectionIII}, the physics-embedded neural dynamics modeling is presented. Section \ref{SectionIV} introduces the detailed deployment on the hardware. The results are provided in Section \ref{SectionV}. The conclusion is given in Section \ref{SectionVI}.

\section{Event Automata Learning Framework for Multi-modal Hybrid Systems} \label{SectionII}

\subsection{Formalization of Multi-Modal PES}

\begin{figure}[t] 
    \centering
    \vspace{-10pt}
    \includegraphics[width=0.48\textwidth]{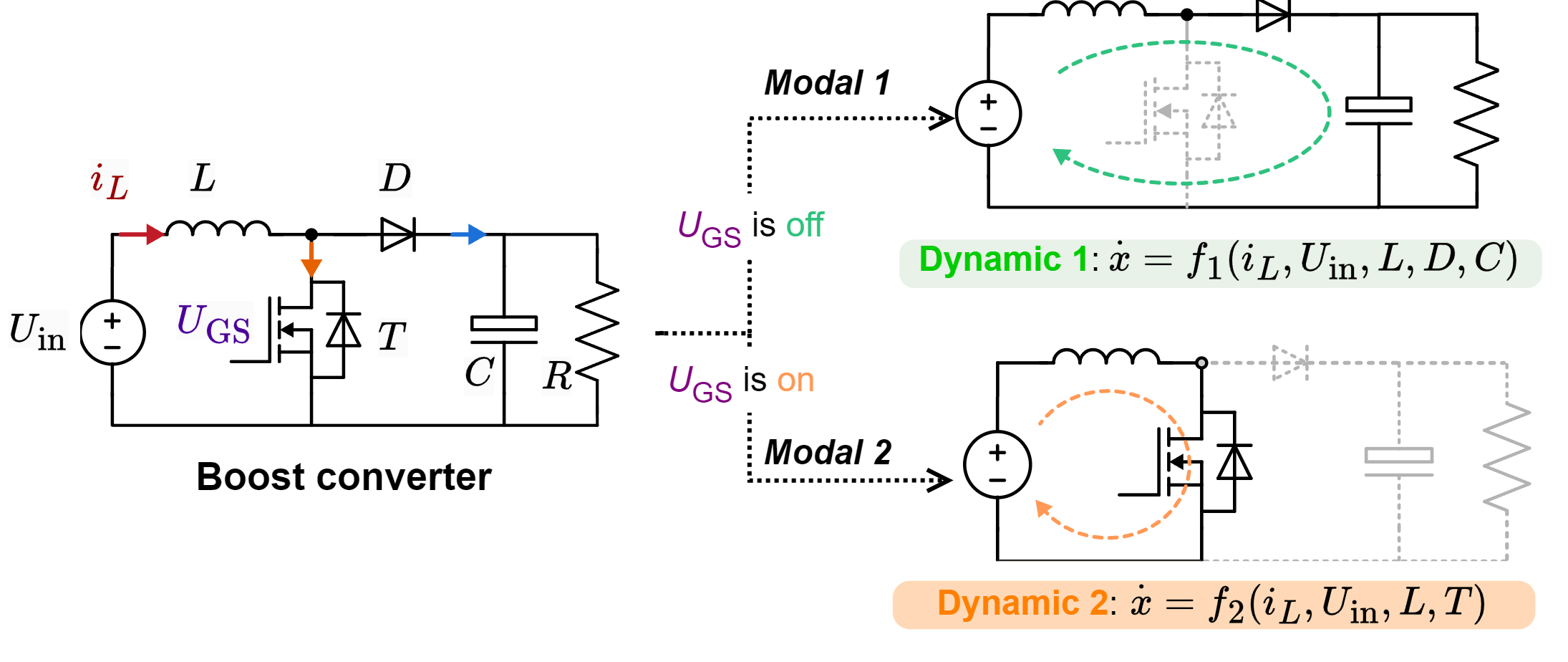} 
    \vspace{-15pt}
    \caption{Multi-modal dynamics of boost converters.}
    \label{boost}
\end{figure}

\begin{figure}[t] 
    \centering
    \vspace{-15pt}
    \includegraphics[width=0.5\textwidth]{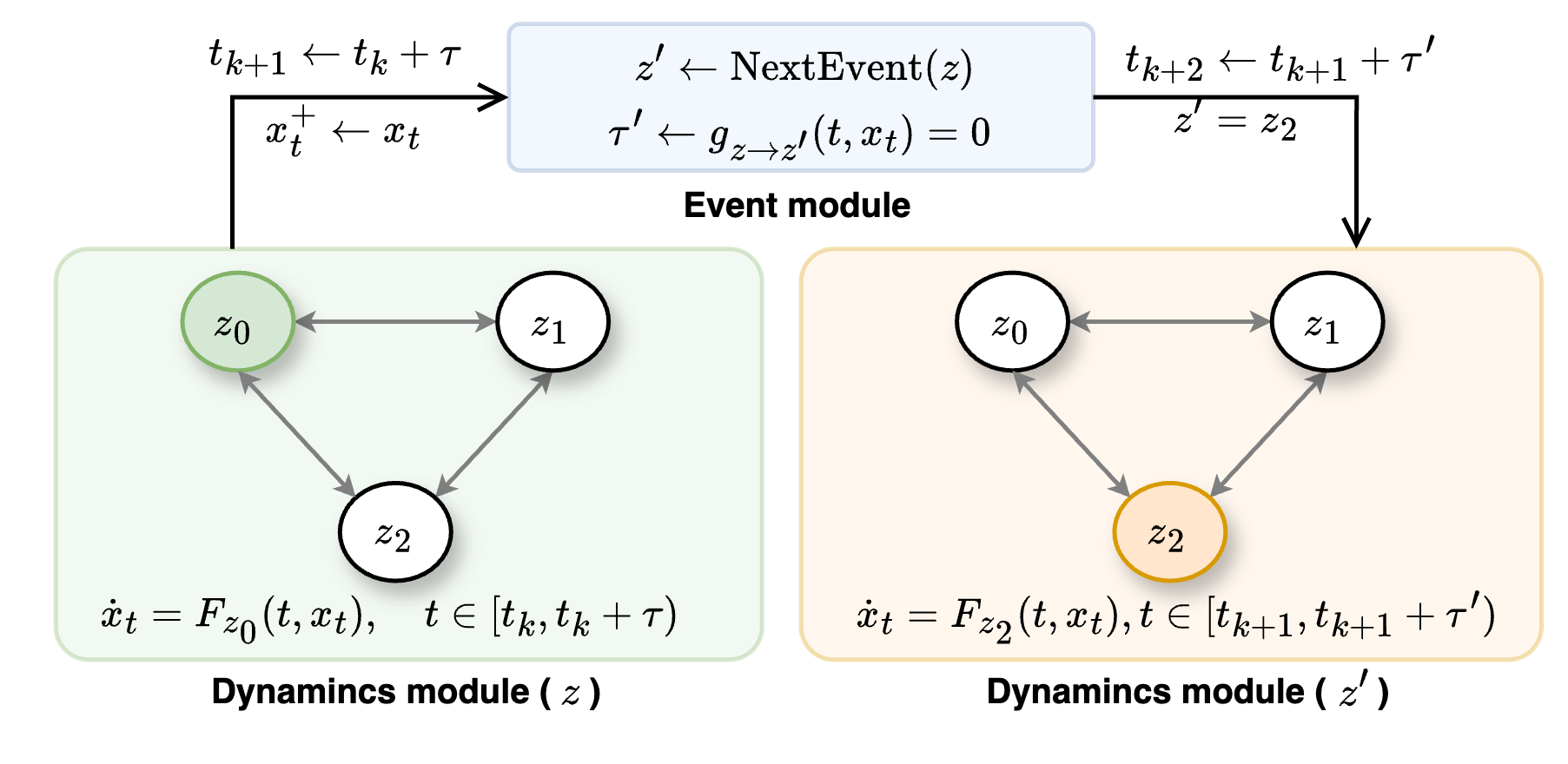} 
    \vspace{-25pt}
    \caption{Mathematical formalization of multi-modal PES.}
    \vspace{-10pt}
    \label{Formalization}
\end{figure}

\begin{figure*}[t]
\begin{minipage}{\textwidth}
\centering
\vspace{-0.5cm}
\includegraphics[width=1\textwidth]{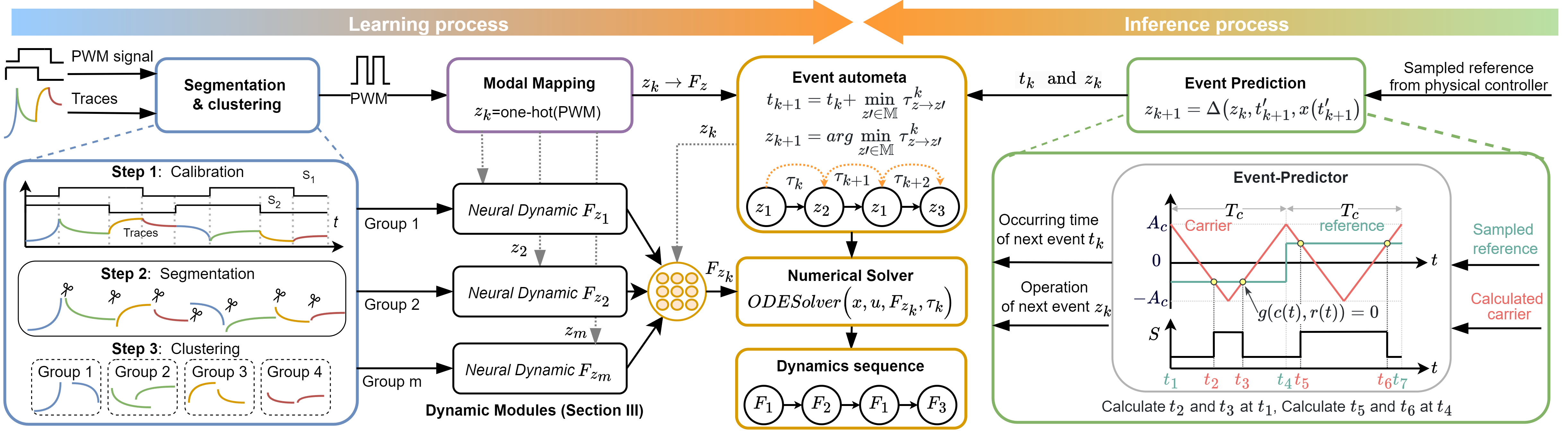}
\vspace{-0.7cm}
\caption{Event-Automata Learning Architecture.}
\label{fig:Event-automata Framework}
\end{minipage}
\vspace{-0.6cm}
\end{figure*}

The operation of PES generally relies on discrete switching actions of power semiconductors, which causes multi-modal dynamics. Take a boost converter shown in Fig. \ref{boost} as an example. The switching device leads to two circuit topologies. In the switching-\textsc{on} state, the inductor current goes through the switch path while the diode path during the switching-\textsc{off} state. The system switches alternately between these modes at a switching frequency from kHz to MHz, causing a piecewise continuous vector field with discontinuous boundaries.


Therefore, it is difficult for data-driven methods to learn transient characteristics between multiple modes, which requires large Neural Networks (NNs) for the system to coordinate different vector fields and approximate discontinuous phase space trajectories. Multiple attractors in a single NN will also cause interference in the backpropagation, i.e., the gradient update for a mode will destroy the representation of other modes. Model dimensionality has to be increased to maintain accuracy, leading to a simultaneous expansion in parameter numbers and training data requirements.

To fill the gap, it is necessary to build a rigorous mathematical formalization to analyze multi-modal PES, as is shown in Fig. \ref{Formalization}. It contains a event module and several dynamics module to represent the synergy of continuous state evolution and discrete events in PES operation, which are denoted as  $x(t)$ and $z(t)$ ($z=1$ and $z=2$ for the switch \textsc{on} and \textsc{off}).   
\begin{align}
  x(t)\in\mathcal{X}\subseteq\mathbb{R}^{n_x}, 
  \qquad
  z(t)\in\mathcal{Z}:=\{1,\dots,Z\}.
\label{ordinary function}
\end{align} 

During a event \(z\), the state obeys its specific ODE.
\begin{equation}
  \dot{x}(t)=F_{z}\!\bigl(t,x(t)\bigr),
  \qquad
  (z,t,x)\in\mathcal{Z}\times\mathcal{T}\times\mathcal{X}.
  \label{eq:mode_ode}
\end{equation}

\noindent where \(F_z\) is the dynamics vector field in a single modal. \(x(t)\) evolves according to \eqref{eq:mode_ode} until another event \(z'\in\mathcal{Z}\) occurs, which triggers dynamics from \(F_{z}\) to \(F_{z'}\). For example, the boost converter has modal dynamics as ${{F}_{z=1}}$ and ${{F}_{z=2}}$ corresponding to the switch \textsc{on} and \textsc{off}. The transition condition can be formalized as a scalar function \(g:\mathcal{T}\times\mathcal{X}\to\mathbb{R}\).
\begin{equation}
  g\!\bigl(t_{k},x(t_{k})\bigr)=0
  \;\;\Longleftrightarrow\;\;
  \text{event at }t_{k}
  \label{eq:event_condition}
\end{equation}

Define a jump set $\mathcal{E}$ to represent the jump conditions between two modes.
\begin{equation}
  \mathcal{E}:=\{(t,x)\mid g(t,x)=0\}.
  \label{eq:event_condition_set}
\end{equation}

Thus, the PES operates with event $z_k$ and modal dynamics \(F_{z_k}\) and evolves continuously until the next event  \(z_{k+1}\) occurs at time \(t_{k+1}\) , which yields the dwell time
\(\tau_k = t_{k+1}-t_k\) and the next state \(x(t_{k+1})\). 
\(z_{k+1}\) can be determined by the transition map $\Delta:\mathcal{Z}\times\mathcal{E}\to\mathcal{Z}$ with \(z_k\) and \(\mathcal{E}\).
\begin{align}
  z_{k+1} \;=\;
  \Delta\!\bigl(z_k, t_{k+1}, x(t_{k+1})\bigr).
\end{align}

Simultaneously, the dynamics during \(z_{k+1}\) are updated to the new vector field \(F_{z_{k+1}}\).
Note that \(x(t_{k+1}^{'})\) remains unchanged during the jump, which provides the initial condition for the subsequent
evolution under \(F_{z_{k+1}}\). In the model learning and inference process, discrete events should be identified and predicted to achieve efficient multi-modal PES modeling.

\subsection{Event-Automata Learning Architecture}

An event-automata learning architecture is proposed in Fig. \ref{fig:Event-automata Framework} to model multi-modal dynamics via identification and characterization of individual modes. In this framework, the modeling is divided into two main process: learning process that learns the PES dynamics through historical data and inference process that calculates the dynamics trajectory based on new real-time signals. 

In the learning phase, complete historical data of control signals are used for offline encoding. The event automata parses the signals into a modal sequence $\left\{z_k\right\}$ through one-hot, obtains the corresponding switching time sequence $\left\{t_k\right\}$, and segments and clusters the original trajectory based on the modal characteristics. Voltage and current trajectories are thus allocated to corresponding dynamic modules, which achieves accurate pattern matching and training.

In the inference phase, control signals are no longer given in advance, but are updated in real time. The framework analyzes the control signal generation mechanism to achieve efficient event prediction, thus avoiding sampling delays, reduced accuracy and massive data introduced by high-frequency sampling methods. Taking the typical PWM control as an example, the switching event is triggered by periodic comparisons between the carrier $c(t)$ and the reference $r(t)$:
\begin{align}
g\left(  t  \right)=c\left( t \right)-r\left( t \right)=0
\label{ordinary function}
\end{align}

The carrier typically adopts sawtooth waveform:
\begin{align}
c(t)=2\cdot{{A}_{c}}\cdot \text{mod}\left( {t}/{{{T}_{c}}}\;,1 \right)-{{A}_{c}}
\label{ordinary function}
\end{align}
where $A_c$ is the magnitude and $T_c$ is the period.

By sampling $r(t)$ and solving (3), the control signals within a switching cycle can be predicted and then one-hot encoded during training to determine the potential modes. The event time sequence $\left\{ t_k \right\}$ and the mode jump path $z\to {z}'$ can also be explicitly obtained. The set of events in each cycle $t\in ({{t}_{k}},{{t}_{k}}+{{T}_{c}})$ can be expressed as:
\begin{align}
   t_{k+1} &= t_k + \min_{z' \in \mathbb{M}} \tau_{z \to z'}^k, \quad z_{k+1} = \operatorname*{arg\,min}_{z' \in \mathbb{M}} \tau_{z \to z'}^k
\label{ordinary function}
\end{align}	
where $\tau _{z\to {z}'}^{k}={{t}_{k+1}}-{{t}_{k}}$ is the modal duration.

Note that the earliest event is selected to determine the next modal and no other events will occur between $[t_k, t_{k+1})$. At $t_{k+1}$, the event automata will save the potential state information for the dynamic module of next modal.
With identified $[t_k, t_{k+1})$ for each mode, the event automata selects ${{\dot{x}}_{t}}={{F}_{{{z}_{k}}}}(t,{{x}_{t}})$ corresponding to $z_k$ and uses an ODE solver (e.g., Dopri5 or RK4) for numerical integration to obtain evolution predictions of the overall state of the system.

\section{Physics-Embedded Neural Dynamics Modeling for Seamless Sim2Real Transfer} \label{SectionIII}

\subsection{The Continuous Neuralization of PES Dynamics}

\begin{figure}[t] 
    \centering
    \includegraphics[width=0.46\textwidth]{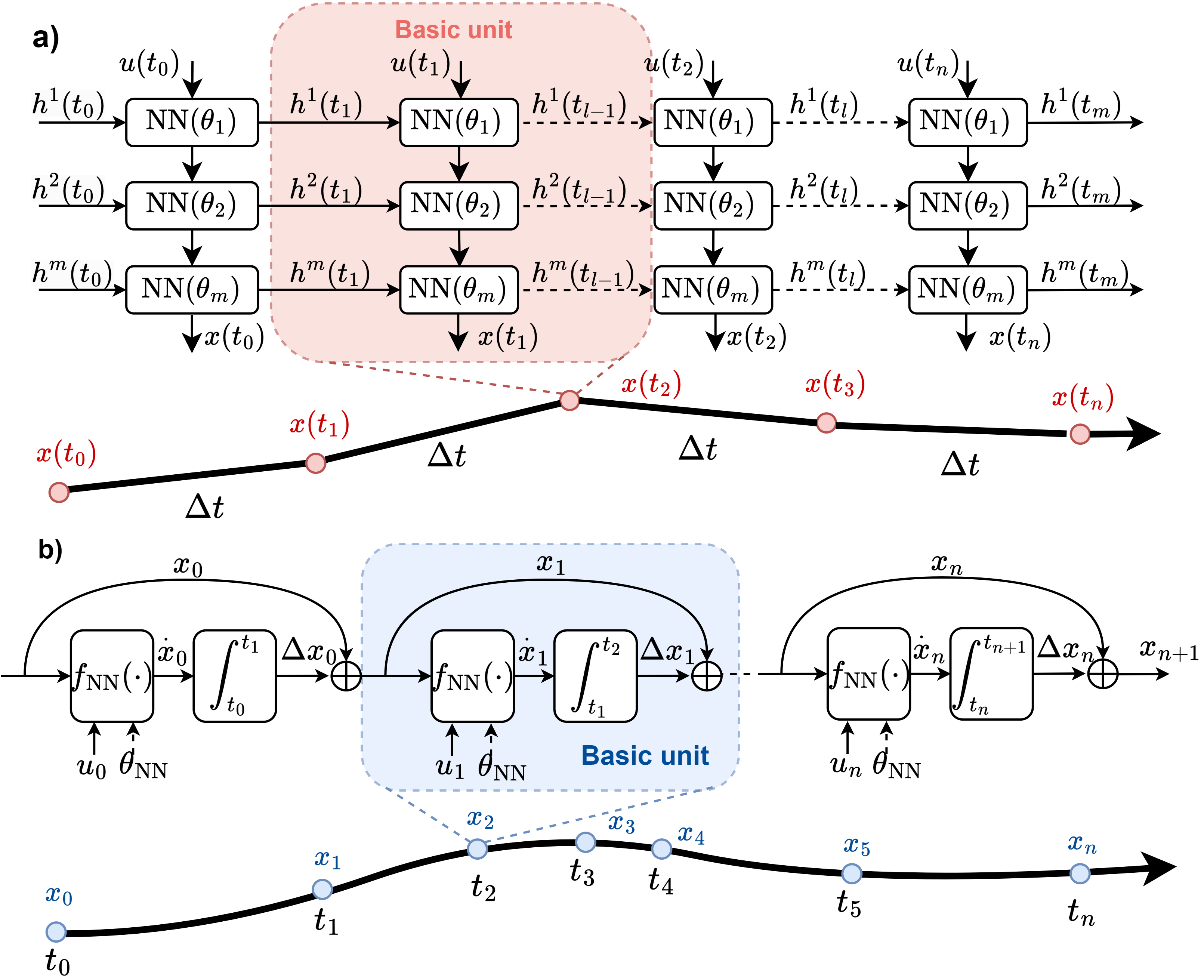} 
    \vspace{-10pt} 
    \caption{Continuous neuralization of dynamics functions. (a).Discrete neural network modeling with m-layer RNN $x_{n+1} = x_{n}+f_{NN}(x_{n}, \theta_{n})$. (b). Continuous neural ODE network modeling $\frac{dx}{dt}=f_{NN}(x, \theta)$.}
    \label{Continuous Neutralize}
    \vspace{-10pt} 
\end{figure}

The event automaton partitions system operation into a set of discrete modes and the continuous time dynamics of each mode must be modeled independently. However, fixed time-step discrete neural networks (RNNs, LSTMs, GRUs) are limited in learning the dynamics of PES. As is shown in Fig. \ref{Continuous Neutralize}(a), Their enforced uniform temporal alignment fractures the inherently continuous and non uniformly switching temporal evolution, making instantaneous cross mode transitions and fast transients difficult to capture faithfully.

As known, the continuous-time dynamics of each hybrid operating mode \( z \in \mathcal{Z} \) in PES satisfies,
\begin{align}
\dot{x}\left( t \right) = f_{z}(x\left( t \right),u\left( t \right)).
\label{Eq:original ordinary function}
\end{align}	
To avoid the drawbacks of discrete neural network, we directly parameterize the continuous-time governing ODE with a neural network, yielding a continuous-time Neural ODE (NODE),
\begin{align}
\dot{x}\left( t \right)={{f}_{NN}}(x(t),u(t), \theta _{\text{NN}}),
\label{ordinary function}
\end{align}	
where ${{f}_{NN}}$ is a neural network (NN) parameterization of the ODE vector field in Eq.~\eqref{Eq:original ordinary function}, $ u(t)$ is the system input vectors and ${{\theta }_{\text{NN}}}$ are the trainable  parameters.  As is shown in Fig. \ref{Continuous Neutralize}(b), this NN model integrates this vector field from \(x(t_{0})=x_{0}\) to obtain the continuous-time trajectory \(x(t)\).
\begin{align}
x(t)= x({{t}_{0}})+\int_{{{t}_{0}}}^{t}{{{f}_{NN}}}(x(\tau ),u(\tau ))d\tau, \ \ \ x({{t}_{0}})={{x}_{0}}
\label{ordinary function}
\end{align}	

The numerical integration with adaptive step size improves the modeling efficiency. The step size is increased when the state is relatively stable and reduced when the state suddenly changes. The final state can be expressed as
\begin{align}
x(t)=\text{ODESolve}\left( {{f}_{NN}},{{x}_{0}},[{{t}_{0}},t] \right)
\label{ordinary function}
\end{align}	

In addition, NODE can directly infer system trajectories through numerical integrators and accurately control error accumulation, making it more flexible and efficient in modeling PES with different modal durations and state change rates.

\subsection{Physics-Embedded Neural Dynamics Structure}

Although neural ODEs are capable of continuous-time modeling, they require large amounts of data to capture dynamic laws. In industry, a wealth of prior physical knowledge (e.g., circuit equations and device models) has already been embedded in simulation tools (e.g., Simulink and PLECS) as ODEs, which is naturally compatible with neural ODEs. This paper thus proposes a Physics-Embedded Neural ODES (PENODE) based on the multi-modal event automata, which achieves explicit physical constraints by integrating known physical laws with data-driven modeling. Specifically, PENODE decomposes the dynamics of one mode into a physical prior term $f_{phy}$ and an NN residual term $f_{NN}$, as is shown in Fig. \ref{Physics Embedded}. (Mode $z$ is omitted in the following analysis):
\begin{align}
\frac{d\mathbf{x}}{dt}={{f}_{\text{phy}}}(\mathbf{x},\mathbf{u},\theta_{\text{phy}}) +{{f}_{\text{NN}}}(\mathbf{x},\theta _{\text{NN}}),\ t\in [{{t}_{k}},{{t}_{k+1}})
\label{ordinary function}
\end{align}		
where ${{\theta }_{\text{phy}}}\in {{\mathbb{R}}^{p}}$ represents physical intrinsic parameters (e.g., $L$, $C$, $R$) and  the $f_{\text{phy}}$ can be expressed from Eq.\ref{Eq:original ordinary function} as,
\begin{align}
{{f}_{\text{phy}}}(\mathbf{x}(t),t,\mathbf{u}(t),{{\theta }_{\text{phy}}})=Ax(t)+Bu(t)
\label{ordinary function}
\end{align}		

The linear form of ${{f}_{\text{phy}}}$ is consistent with the local continuous system while ${{f}_{NN}}$ approximates degradation modeling or environmental effects. Since neural ODEs guarantee the time dependency, a multi-layer perceptron (MLP) can be used to model the dynamics. Its parameters ${{\theta }_{\text{NN}}}$ includes weights ${{\mathbf{W}}^{(j)}}$ and biases ${{\mathbf{b}}^{(j)}}$. It is considered that for $nx$ -dimensional state variables, the perfect form for any dimensional known physical term and neural network term can be expressed as,

\begin{align}
\frac{d\mathbf{x}(t)}{dt}=\left[ \begin{matrix}
   {{\mathbf{0}}_{{{p}_{1}}\times 1}}  \\
   {{f}_{\text{phy}}}(\mathbf{x}(t),t,\mathbf{u}(t),{{\theta }_{phy}})  \\
   {{\mathbf{0}}_{{{p}_{2}}\times 1}}  \\
\end{matrix} \right]+\left[ \begin{matrix}
   {{\mathbf{0}}_{{{p}_{3}}\times 1}}  \\
   {{f}_{\text{NN}}}(\mathbf{x}(t),t,\text{ }{{\theta }_{NN}})  \\
   {{\mathbf{0}}_{{{p}_{4}}\times 1}}  \\
\end{matrix} \right]
\label{ordinary function}
\end{align}			
where ${{\mathbf{0}}_{{{p}_{1}}\times 1}}$, ${{\mathbf{0}}_{{{p}_{2}}\times 1}}$, ${{\mathbf{0}}_{{{p}_{3}}\times 1}}$and ${{\mathbf{0}}_{{{p}_{4}}\times 1}}$ are zero vectors, allowing flexible physical constraints and capturing nonlinear residual state variables.

\subsection{Unified  Parameter Learning Strategy}

\begin{figure}[t] 
    \centering
    \vspace{-15pt} 
    \includegraphics[width=0.5\textwidth]{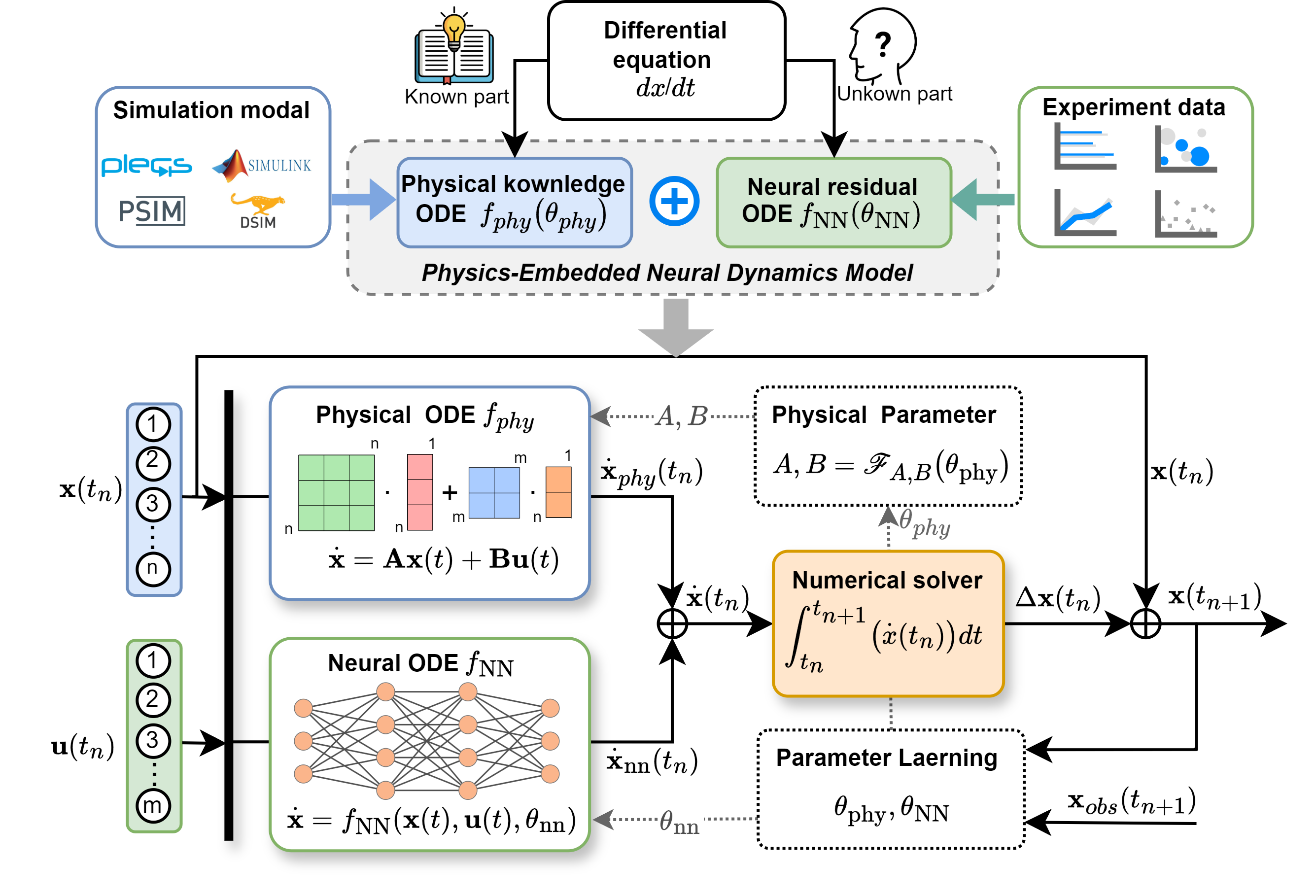} 
    \vspace{-20pt} 
    \caption{Physics-embedded neural ODEs with unified parameter learning strategy.}
    \vspace{-10pt} 
    \label{Physics Embedded}
\end{figure}

PENODE achieves optimization of physical parameters and NN weights through forward and backward propagation (i.e., forward state evolution and reverse gradient propagation), which synchronously update heterogeneous parameters under a unified architecture, as shown in Fig. \ref{Physics Embedded}.

\textbf{Forward Propagation:} given an initial parameter set $\Phi =\{{{\theta }_{\text{phy}}},{{\theta }_{\text{NN}}}\}$, the system matrix $A$ and the control matrix $B$ are explicitly constructed from it.
\begin{align}
A={{\mathcal{F}}_{A}}({{\theta }_{\text{phy}}})\in {{\mathbb{R}}^{d\times d}},\quad B={{\mathcal{F}}_{B}}({{\theta }_{\text{phy}}})\in {{\mathbb{R}}^{d\times m}}
\label{ordinary function}
\end{align}	
where ${{\mathcal{F}}_{A}}$ and ${{\mathcal{F}}_{B}}$ are matrix constructors based on physical laws. Then, the system dynamics is described by the physically dominant term and the NN residual term.
\begin{align}
\dot{x}=\underbrace{Ax+Bu}_{Dominant}+\underbrace{{{f}_{\text{NN}}}(x,t,{{\theta }_{\text{NN}}})}_{Residual}
\label{ordinary function}
\end{align}	
	
${{\theta }_{\text{phy}}}$ is initialized to the nominal values of offline simulation $\theta _{\text{phy}}^{\text{sim}}={{[{{L}_{\text{sim}}},{{C}_{\text{sim}}},{{R}_{\text{sim}}},\cdots ]}^{T}}$  to inherit prior physical knowledge while ${{\theta }_{\text{NN}}}$ is initialized using the He Normal \cite{DBLP:journals/corr/abs-2004-06632}. The output layer weights are initialized by a Gaussian distribution ($\mu =0, \sigma =0.01$), avoiding interference with the initial dynamics.
The numerical integration of the state trajectory sequence $\{x({{t}_{k}})\}_{k=1}^{K}$ is implemented by the Dormand-Prince (RK45) adaptive step solver, which improves computational efficiency while ensuring numerical accuracy.

\textbf{Back Propagation}:  In order to train the model, it is necessary to obtain the observation sequence $\{{{x}_{\text{obs}}}({{t}_{k}})\}_{k=1}^{K}$ from the actual PES and construct a multi-step prediction loss function after wavelet threshold denoising reprocessing.
\begin{align}
    \mathcal{L}_{} &:= 
    \frac{1}{K}\sum_{k=1}^{K}
    \bigl\|x_{\text{}}(t_k)-x_{\text{obs}}(t_k)\bigr\|_2^2. 
\end{align}

To optimize $\Theta = \{\theta_{\text{phy}}, \theta_{\text{NN}}\}$, the gradients of $\mathcal{L}(\Theta)$ with respect to $\Theta$ are needed, which can be computed using the memory-efficient adjoint sensitivity method or automatic differentiation (AD) frameworks that perform backpropagation through ODE solver operations. 

Specifically, an adjoint state $\mathbf{a}_{\mathbf{x}}(t)$ representing the sensitivity of the loss $\mathcal{L}$ to $\mathbf{x}(t)$ is propagated backward.
\begin{align}
\frac{d\mathbf{a}_{\mathbf{x}}(t)}{dt} = -\mathbf{a}_{\mathbf{x}}(t)^T \left( \frac{\partial f_{\text{phy}}}{\partial \mathbf{x}} + \frac{\partial f_{\text{NN}}}{\partial \mathbf{x}} \right) \label{eq:adjoint_dynamics}
\end{align}

The backward propagation starts with $\mathbf{a}_{\mathbf{x}}(t_K) = \nabla_{\mathbf{x}(t_K)} \mathcal{L}$, and contributions of $\mathcal{L}$ are added at each $t_k$.
The gradients of $\mathcal{L}$ with respect to any $\theta \in \Theta$ are then computed as:
\begin{align}
\frac{\partial \mathcal{L}}{\partial \theta} = \int_{t_0}^{t_K} \mathbf{a}_{\mathbf{x}}(\tau)^T \frac{\partial (f_{\text{phy}}(\theta_{\text{phy}}) + f_{\text{NN}}(\theta_{\text{NN}}))}{\partial \theta} d\tau \label{eq:gradient_data_term}
\end{align}

Finally, the gradient is then supplied to an Optimizer, such as Adam, to update parameters $\Theta$ iteratively using the Adam optimizer as follows:
\begin{align}
\Theta^{(k+1)} = \Theta^{(k)} - \eta \cdot \mathrm{Adam}\left( \nabla_{\Theta} \mathcal{L}(\Theta^{(k)}) \right)
\end{align}
where $\eta$ is the learning rate, and $\nabla_{\Theta} \mathcal{L}$ denotes the gradient of the loss with respect to all parameters.

\subsection{Sim-to-Real Transferability}

The proposed PENODE architecture initializes with known ($\theta_{\text{phy}}^{\text{sim}}$) from a simulator and fine-tunes both $\theta_{\text{phy}}$ and $\theta_{\text{NN}}$ using real-world data, which can be used for robust Sim-to-Real transfer in diverse scenarios, including white-box, gray-box, and black-box modeling contexts.

\begin{itemize}
   \item \textbf{White-box scenarios:} If physical models are well-known and accurate, $f_{\text{NN}}$ can be minimal or even zero, and the learning refines $\theta_{\text{phy}}$ to match the real system.
   \item \textbf{Gray-box scenarios:} When only a partial physical model is available, $f_{\text{phy}}$ captures the known dynamics, while $f_{\text{NN}}$ learns the unknown or poorly modeled parts. This is the most common scenario for PENODE.
   \item \textbf{Black-box scenarios:} If no prior physical knowledge is available, $f_{\text{phy}}$ can be set to zero (or a very generic form), and $f_{\text{NN}}$ learns the entire dynamics from data, behaving like a standard Neural ODE.
\end{itemize}
This adaptability with the progressive learning strategy ensures that PENODE can effectively transfer from simulations to real-world hardware with high fidelity, even with limited real-world data. The explicit parameterization of physical components also aids in interpretability, as changes in $\theta_{\text{phy}}$ can often be related back to physical changes in the system.

\begin{figure}[t] 
    \centering
    \vspace{-15pt} 
    \includegraphics[width=0.46\textwidth]{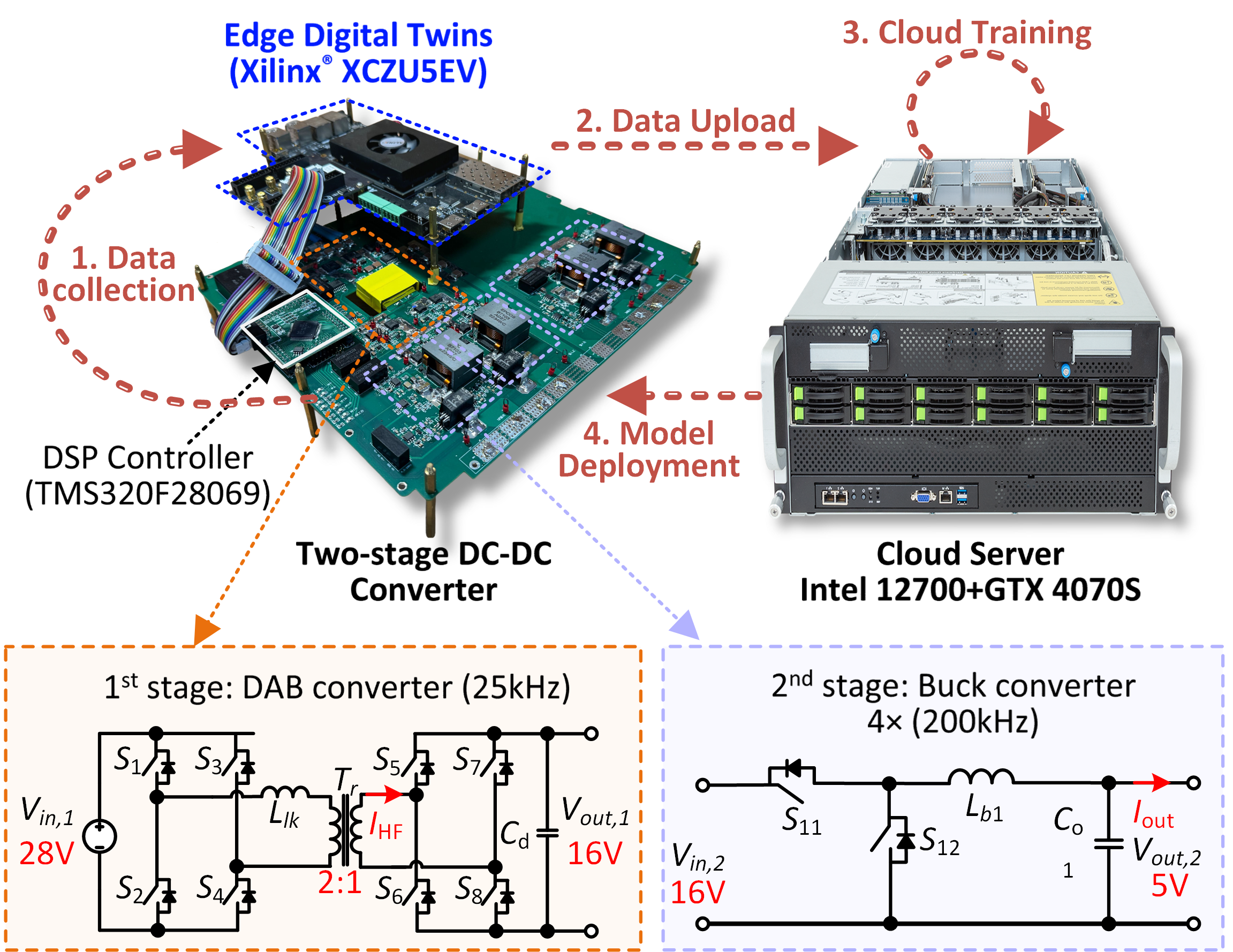} 
    \vspace{-8pt} 
    \caption{ EDT Hardware platform and the studied converter topologies.}
    \vspace{-8pt} 
    \label{fig: platform}
\end{figure}

\section{Design, Training and Implemention of PENODE-based Edge Digital Twin }  \label{SectionIV}


\subsection{Edge Digital Twin Platform of Studied Case}

\begin{figure*}[t]
\begin{minipage}{\textwidth}
\centering
\vspace{-0.5cm}
\includegraphics[width=1\textwidth]{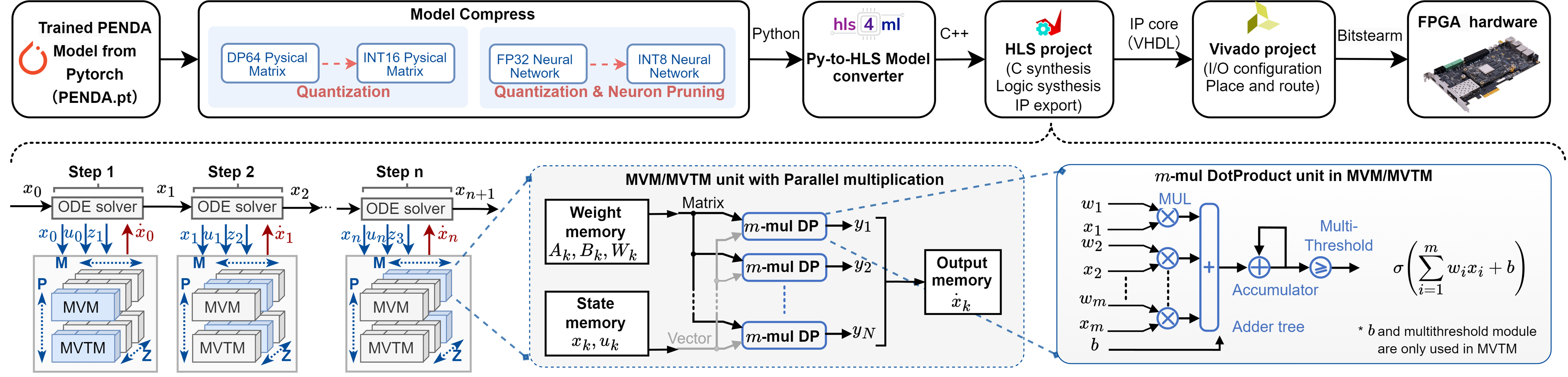}
\vspace{-0.7cm}
\caption{Deployment Workflow of the PENODE Model on FPGA-Based Edge Devices.}
\label{fig:Edge Deployment}
\end{minipage}
\vspace{-0.5cm}
\end{figure*}

A heterogeneous edge digital twin platform is designed and constructed, as illustrated in Fig. \ref{fig: platform} which comprises a cloud computing platform for PENODE model training, an edge platform for PENODE model inference, and a physical converter under study. The studied case is a 600 W two-stage DC-DC converter with a 25 kHz DAB and four parallel 200 kHz buck converters, all controlled by TI TMS320F28069, whose parameters are listed in Table I. The edge platform includes a low-cost Xilinx Zynq UltraScale+ MPSoC development board (XCZU5EV) while the cloud platform utilizes an NVIDIA GeForce GTX 4070 Super GPU and an Intel Core i7-12700 CPU for PENODE model training. 

The EDT can readily acquire the state variables by voltage/current sensors and control signals through ADCs and transmit them to the cloud for model training. The trained model is deployed onto the edge device for real-time inference. The EDT can also modify the control parameters to obtain data under various conditions and capture different operating modes. In this study, 1200 sets of data were collected under different control parameters, stored locally on the edge device, and uploaded to the cloud by TCP/IP over a LAN interface.

\begin{table}[!t]
  \centering

  \caption{Design Parameters of the two-stage converter}
  \label{tab:twostage_params}
  \renewcommand{\arraystretch}{1.0}
  \begin{tabular}{>{\centering\arraybackslash}m{2.4cm}
                  >{\centering\arraybackslash}m{1.2cm}
                  >{\centering\arraybackslash}m{2.5cm}
                  >{\centering\arraybackslash}m{1.2cm}}
    \toprule
    \textbf{Variable} & \textbf{Value} & \textbf{Variable} & \textbf{Value} \\ \midrule
   Input voltage $V_{\text{in,1}}$      & 16--50 V  & Output voltage $V_{out}$& 5 V\\
    Input-port power &  600 W & Output-port power                       & 150 W \\
    DAB frequency & 25 kHz & Buck frequency& 200 kHz\\
    Bus Capacitance $C_{d}$& 1.76 mF& DC Capacitance $C_{o1}$& 94 $\mu$F\\
    DAB inductance $L_\text{lk}$ & 2 µH & Buck inductance $L_\text{b}$ & 10 µH \\                     
    \bottomrule
  \end{tabular}
  \vspace{-5pt}
  \begin{flushleft}
    \footnotesize * Switches $S_1$–$S_8$ in DAB are IPT017N12NM62 . **Switches$S_{11}$, $S_{12}$, $S_{21}$, $S_{22}$ $S_{31}$, $S_{32}$ $S_{41}$, $S_{42}$ in Buck are IAUA250N04S6N005. 
  \end{flushleft}
  \vspace{-18pt}
\end{table}

\subsection{PENODE Model Training on Cloud}

\begin{algorithm}[t]
\caption{End-to-End Training Pipeline of PENODE}
\label{training_pipeline}
\begin{algorithmic}[1]
\Require Raw dataset $D$; physical template $\mathcal{M}_{\text{phy}}$;  
         loss $\mathcal{L}$ ; optimizer $\mathcal{O}$ ;  
         max iterations $K$; tolerance $\varepsilon$
\Ensure  Trained parameters $\theta_{\text{phy}}^{\star},\;\theta_{\text{nn}}^{\star}$

\Statex \textbf{Step 0: Data Pre‐processing}
\State Align timestamps and apply sliding‐window denoising
\State Cluster samples by switching combination
\State Split into $\mathcal{D}_{\text{train}},\;\mathcal{D}_{\text{val}},\;\mathcal{D}_{\text{test}}$

\Statex \textbf{Step 1: Physical Term Pre‐training}
\State Build simulation model $\mathcal{M}_{\text{phy}}$ from prior methodology
\State Initialize $\theta_{\text{phy}}$
\While{not converged}
    \State $y \gets \texttt{ODESolve}\bigl(f_{\text{phy}},\theta_{\text{phy}},\mathcal{D}_{\text{train}}\bigr)$
    \State $\ell \gets \mathcal{L}\!\left(y,\mathcal{D}_{\text{train}}\right)$
    \State $\theta_{\text{phy}} \gets \mathcal{O}\!\left(\theta_{\text{phy}},\ell\right)$
\EndWhile
\State $\theta_{\text{phy}}^{\text{best}} \gets \arg\min \ell$

\Statex \textbf{Step 2: Hyperparameter Optimization}
\State Use Bayesian optimization to tune learning rate, batch size, etc.\ for subsequent steps

\Statex \textbf{Step 3: Joint Training of $f_{\text{phy}}$ and $f_{\text{nn}}$}
\State Initialize $\theta_{\text{phy}}\!\gets\!\theta_{\text{phy}}^{\text{best}}$, \;randomly initialize $\theta_{\text{nn}}$
\While{not converged}
    \State $y_{\text{phy}} \gets \texttt{ODESolve}\bigl(f_{\text{phy}},\theta_{\text{phy}},\mathcal{D}_{\text{train}}\bigr)$
    \State $y_{\text{nn}}  \gets \texttt{DiffODESolve}\bigl(f_{\text{nn}},\theta_{\text{nn}},\mathcal{D}_{\text{train}}\bigr)$
    \State $\ell \gets \mathcal{L}\!\left(y_{\text{phy}}+y_{\text{nn}},\mathcal{D}_{\text{train}}\right)$
    \State $(\theta_{\text{phy}},\theta_{\text{nn}}) \gets \mathcal{O}\!\bigl((\theta_{\text{phy}},\theta_{\text{nn}}),\ell\bigr)$
\EndWhile
\State $(\theta_{\text{phy}}^{\star},\theta_{\text{nn}}^{\star}) \gets \arg\min \ell$

\Statex \textbf{Step 4: Iterative Refinement}
\State Repeat \textbf{Steps 2–3} until $|\Delta\ell|<\varepsilon$ or iteration $=K$

\Statex \textbf{Step 5: Model Export}
\State Validate on $\mathcal{D}_{\text{val}}$
\State Save $(\theta_{\text{phy}}^{\star},\theta_{\text{nn}}^{\star})$ to cloud storage
\end{algorithmic}
\end{algorithm}

The training of the PENODE model is on the cloud within a Python-based environment (e.g., the PyTorch deep learning framework). 
The complete training procedure for PENODE is depicted in Algorithm \ref{training_pipeline}.

\subsection{PENODE Model Implemention on Edge Device}

Migrating the cloud-trained model to the edge device requires model compression and optimization to adapt to the hardware constraints of the edge device, as shown in Fig. \ref{fig:Edge Deployment}. 

An open-source PyTorch library Brevitas is used for quantization-aware training. First, the NN residual term undergoes magnitude-based pruning to remove connections with low contributions, resulting in a sparser network structure. Then the coefficients of the linear physical term (matrices $A$ and $B$), which represent core dynamic laws, are mapped from 64-bit floating-point parameters to a 32-bit fixed-point representation. For the residual network, input activations and weights are linearly mapped to an 8-bit integer space, while the output layer retains FP16 precision to mitigate error accumulation. 

Furthermore, HLS4ML is employed to transform the NN model into the edge FPGA, which takes the PyTorch model as input and generates a hardware accelerator. The accelerator is specified in C/C++ and can be synthesized into RTL using Xilinx Vivado HLS. 
Then multi-dimensional parallelization is used in the hardware implementation with $P$ parallel Matrix-Vector Multiply (MVM) units for the physical term and Matrix-Vector Threshold Units (MVTU) for the neural network term. MVMs and MVTUs transform an $m \times n$ matrix and $n$-vector multiplication into $m$ parallel $n$-dimensional vector dot product operations where $n$ vector element multiplications can be performed in parallel. MVTUs additionally perform a "multi-threshold" operation representing the activation function. A Look-Up Table (LUT)-based event automaton is then implemented, which acquires PWM signals and directly maps them to the corresponding modal index. Finally, the C/C++ code generated by HLS4ML is synthesized into an IP core targeting the FPGA using Xilinx Vivado HLS. This IP core is integrated into a Vivado project, synthesized into a bitstream file, and downloaded to the edge device.

\section{Results Analysis and Performance Evaluation}\label{SectionV}

\subsection{Evaluation Scenarios Setup}

\begin{table}[!t]
  \centering
  \vspace{-10pt}
  \scriptsize
  \caption{Training Parameters and Hyper-parameter Search Space}
  \vspace{-5pt}
  \label{tab:train_hparam}
  \renewcommand{\arraystretch}{1.1}
  \begin{tabular}{>{\centering\arraybackslash}m{1.5cm} >{\centering\arraybackslash}m{2.0cm} >{\centering\arraybackslash}m{1.5cm} >{\centering\arraybackslash}m{2.0cm}}
  
    \toprule
    \multicolumn{2}{c|}{\textit{Fixed Training Parameters}} & \multicolumn{2}{c}{\textit{Hyper‑parameter Search Space}} \\ \midrule
    Max epochs & 100 (early stopping patience = 15)  & Learning rate & LogUniform $(10^{-5},10^{-2})$ \\
    Iteration & 1500 & Batch size & \{16,32,64,128,256\} \\
    Optimiser & Adam & \# Layers & 1--4 \\
    Initialisation & He, $[-0.05,0.05]$ & \# Neurons & 16--256 \\
    Activation Function& ReLU & L2 regularization & LogUniform $(10^{-6},10^{-3})$ \\
    \bottomrule
  \end{tabular}
  \vspace{-10pt}
\end{table}

\begin{table}[!t]
  \centering
  \scriptsize
  \renewcommand{\arraystretch}{1.1}
  \setlength{\tabcolsep}{1pt}
  \caption{Evaluation Scenarios Setup}
  \vspace{-5pt}
  \label{tab:evaluation_scenarios}
  \begin{tabular}{>{\centering\arraybackslash}m{1.4cm}
                  >{\centering\arraybackslash}m{2.0cm}
                  >{\centering\arraybackslash}m{3.0cm}
                  >{\centering\arraybackslash}m{2.0cm}}
    \toprule
    \textbf{Scenarios} & \textbf{Learning Efficiency} & \textbf{Sim2Real Transfer Capability} & \textbf{Real-time Edge Performance} \\
    \midrule
    \textbf{Model Type} & Black-box (PENODE-B) & \begin{tabular}[c]{@{}l@{}}Gray-box 1 (PENODE-G1*)\\ Gray-box 2 (PENODE-G2**)\\ White-box (PENODE-W)\end{tabular} & Gray-box (PENODE-G) \\
    \textbf{Training set condition} & $ V_{in} \in [10 V,40 V]$, $ V_{out,1}  = 16 V $ &  $ V_{in}  \in [10 V,40 V]$, $ V_{out,1}  = 16$ &   $ V_{in}  \in [10 V,40 V]$, $ V_{out,1}  = 16 $\\
    \textbf{Dataset Size} & 10\,\%, 25\,\%, 50\,\%, 100\,\% & 10\,\%, 25\,\%, 50\,\% & 10\,\%, 25\,\%, 50\,\% \\
    \textbf{Benchmarks} & \begin{tabular}[c]{@{}l@{}}$\bullet$\;PENODE w/o EA$^{\dag}$\\ $\bullet$\;RNN w/o EA\\ $\bullet$\;LSTM w/o EA\\ \end{tabular} & \begin{tabular}[c]{@{}l@{}}$\bullet$\;Simulation model\\ $\bullet$\;PINN \cite{10463542}\\ $\bullet$\;PRNN\cite{9779551}\end{tabular} & \begin{tabular}[c]{@{}l@{}}$\bullet$\;Fixed-step solver\\ $\bullet$\;Variable-step solver\\ $\bullet$\;Event-driven solver\end{tabular} \\
    \textbf{Generalization test conditions} &   $V_{in} \in [40 V,50 V]$, $V_{out,1} = 24 V $&   $V_{in} \in [40 V,50 V]$, $V_{out,1} = 24 V $&    $V_{in} \in [40 V,50 V]$, $V_{out,1} = 24 V $\\
    \textbf{Evaluation Indicators} & MSE; training hours & MSE; training hours; model size & Inference latency (\textmu s); FPGA utilisation (\%); \\
    \bottomrule
  \end{tabular}
  \vspace{-2pt}
  \begin{flushleft}
    \footnotesize *Gray-box 1: only DAB structure is known; **Gray-box 2: complete structure is known. \quad$^{\dag}$EA: Event automata.
  \end{flushleft}
  \vspace{-20pt}
\end{table}

The PENODE model of the converter was trained offline using a dataset of 1,200 time-series samples, partitioned into training, validation, and test sets in a 7:2:1 ratio. To evaluate robustness under data scarcity, different proportions of data were randomly sampled from the training set for model training. For each proportion, 10 independent trials were run to report the mean and standard deviation of the performance.

The training process utilized the Mean Squared Error (MSE) loss function with the Adam optimizer. A Bayesian optimizer is also deployed to explore an optimal hyperparameter space encompassing key parameters such as learning rate, batch size, and model architecture. Each optimization trial was run for a maximum of 100 epochs with early stopping mechanism. The training parameters and hyperparameter search space are summarized in Table \ref{tab:train_hparam}. To comprehensively evaluate the performance of PENODE, three test scenarios have been designed, as summarized in Table \ref{tab:evaluation_scenarios}.

\begin{figure}[t] 
    \centering
    \vspace{-20pt} 
    \includegraphics[width=0.5\textwidth]{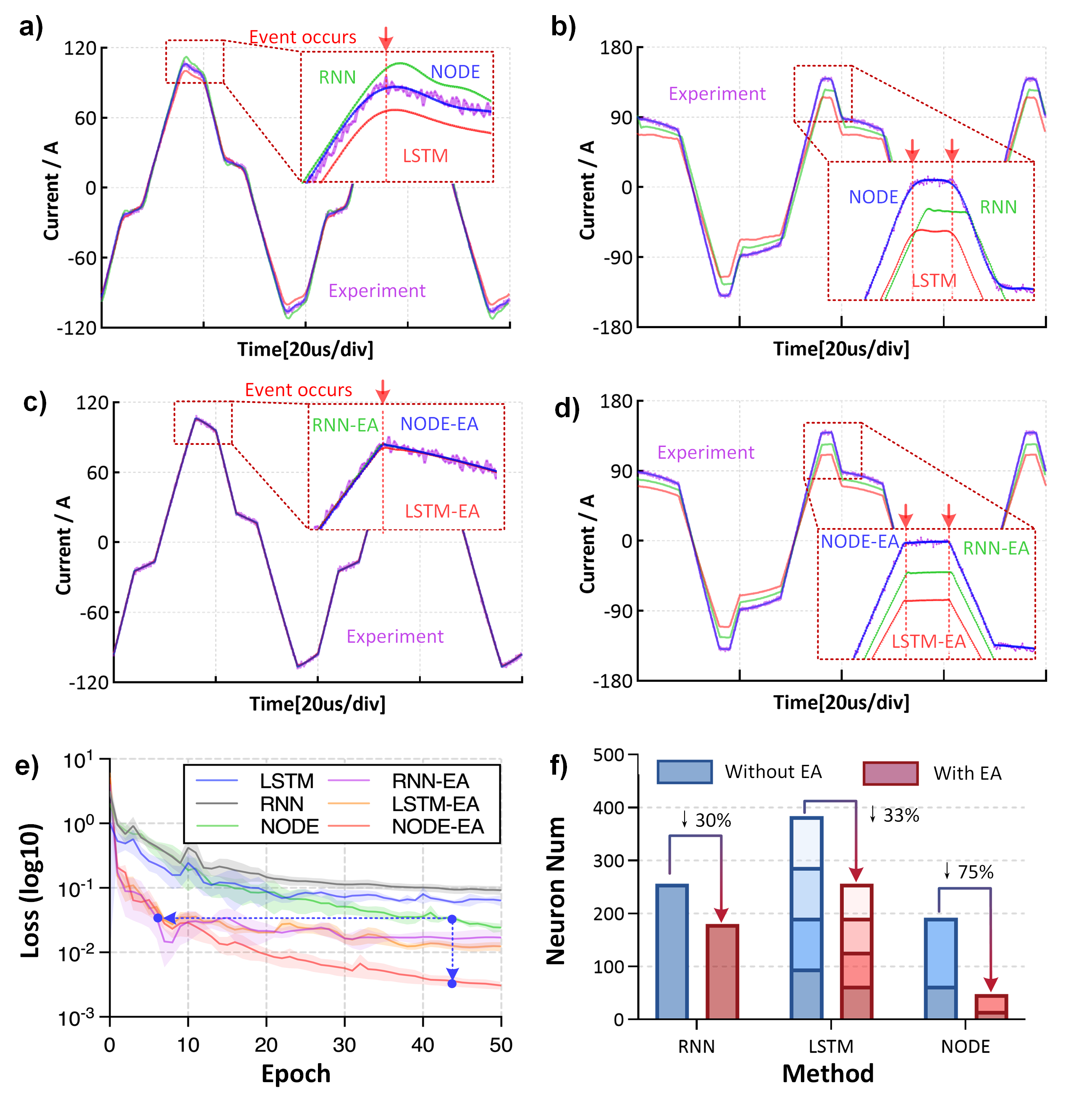} 
    \vspace{-20pt} 
    \caption{Learning Efficiency of the Event-Automata (EA) Framework. (a) In-domain prediction without EA. (b) out-of-domain prediction without EA. (c) in-domain prediction with EA. (d) out-of-domain prediction with EA. (e). Training loss curves for different models. PENODE and EA-based variants converge faster and lower. (f). Model complexity of different methods, including the number of layers and neurons per layer.}
    \label{fig: learning efficiency}
    \vspace{-10pt} 
\end{figure}


\subsection{Learning Efficiency of Event-Automata Framwork}

\begin{figure*}[t]
\begin{minipage}{\textwidth}
\centering
\vspace{-0.5cm}
\includegraphics[width=1\textwidth]{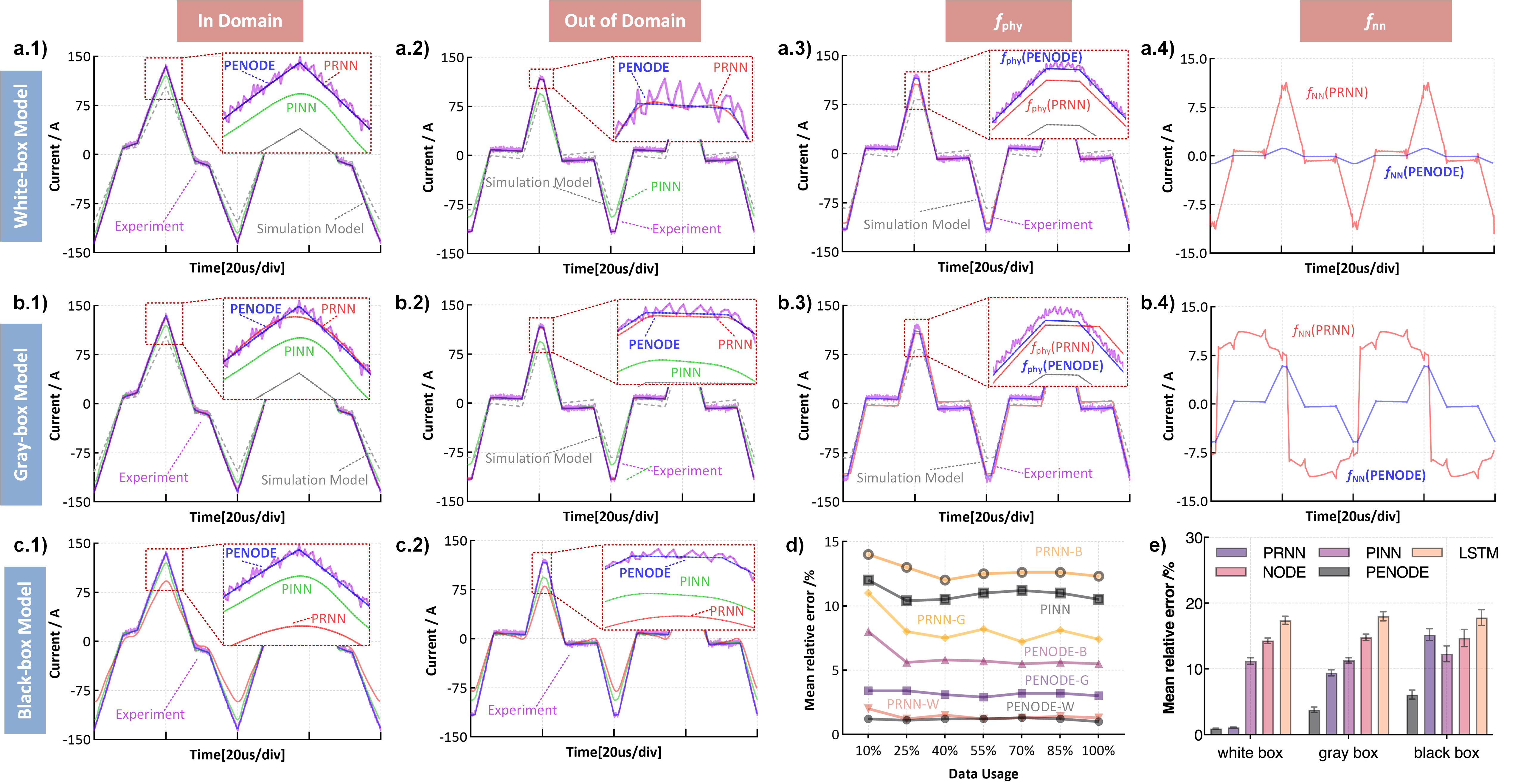}
\vspace{-0.8cm}
\caption{Evaluation of Sim2Real Transfer Capability of PENODE. (a–c). Sim-to-real performance under different model types: white-box (PENODE-W), gray-box (PENODE-G), and black-box (PENODE-B). Subplots *.1 and *.2 show in-domain and out-of-domain current trajectories, respectively; *.3 and *.4 visualize the physical term $f_{\mathrm{phy}}$ and neural residual term $f_{\mathrm{nn}}$. 
    (d). Mean relative error across varying data usage levels for in-domain testing. 
    (e). Out-of-domain mean relative error comparison across model types and methods.}
\vspace{-0.5cm}
\label{fig: test results}
\end{minipage}
\end{figure*}

\begin{table*}[!t]
  \centering
  \footnotesize
  \renewcommand{\arraystretch}{1.1}
  \setlength{\tabcolsep}{3pt}
  \caption{Comparison of PENODE, PRNN, and PINN in Terms of Sim2Real Transfer Capability}
  \vspace{-8pt}
  \label{tab:method_compare_en}
  \begin{tabular}{>{\centering\arraybackslash}p{1.8cm}
                  >{\centering\arraybackslash}p{3.0cm}
                  >{\centering\arraybackslash}p{3.0cm}
                  >{\centering\arraybackslash}p{3.4cm}
                  >{\centering\arraybackslash}p{2.8cm}
                  >{\centering\arraybackslash}p{2.5cm}}
    \toprule
    \textbf{Method} & \textbf{Physics Component} & \textbf{Data‐Driven Residual} & \textbf{Training Paradigm} & \textbf{Traning Solver} & \textbf{Global Framework} \\
    \midrule
    \textbf{PENODE} & Continuous physics ODE $f_{\mathrm{phy}}(x,\theta_{phy})$  & Continious Neural‑ODE residual $f_{\mathrm{NN}}(x,\theta_{NN})$ & End‑to‑end joint optimisation $\theta_{phy} \& \theta_{NN} $ & ODE solver + adjoint method & EA framework for multi-modal systems \\
    \midrule
    \textbf{PRNN}\cite{9779551} & Fixed‑step discretisation $x_{k+1}=x_k + f_{\mathrm{phy}}$ &Discrete residual network $f_{\mathrm{NN}}(x,\theta_{NN}, \Delta t)$   & Two‑stage: (1) fit $f_{\mathrm{phy}}$; (2) freeze $f_{\mathrm{phy}}$, then train $f_{\mathrm{NN}}$ & BPTT in discrete domain &Single-modal framework \\
    \midrule
    \textbf{PINN}\cite{10463542} & Physics laws (KCL/KVL) embedded in loss & $f_{\mathrm{NN}}(x,\theta_{NN})$ directly fitting $x(t)$ & Traning with extra physics‑informed loss & BPTT in discrete domain  & Single-modal framework \\
    \bottomrule
  \end{tabular}
  \vspace{-15pt}
\end{table*}

This scenario evaluates how the proposed Event-Automata framework improves learning efficiency for multi-modal PES. A comparative study is conducted to assess the performance of a discrete NN, RNN and LSTM, and a continuous NN, NODE, both with and without the EA framework on black-box models without any physics-based information. A DAB converter under triple-phase-shift control is tested to highlight the model response to switching events. 50\% of the complete dataset are used for training and a Bayesian optimizer is employed to tune the optimal hyperparameters, with each trial running for a maximum of 50 epochs. After training, the models were statically tested under the in-domain and out-of-domain conditions detailed in Table \ref{tab:evaluation_scenarios}.

As can be seen in Fig. \ref{fig: learning efficiency}(a), NODE matches well with the experimental waveforms without the EA framework and outperforms RNN and LSTM under in-domain conditions. In the out-of-domain tests shown in Fig. \ref{fig: learning efficiency}(b), the performances of RNN and LSTM degrade significantly while NODE maintained robust. However, all these standard neural networks struggled to capture the abrupt dynamic transitions caused by switching events. As shown in Fig. \ref{fig: learning efficiency}(c), the EA framework handling the modal transitions markedly improved the dynamic performance of the networks, especially for RNN and LSTM, allowing the NNs to focus on learning the continuous characteristics within each mode. PENODE still performed best under out-of-domain conditions, as shown in Fig. \ref{fig: learning efficiency}(d). The fundamental reason is that PENODE, through its EA framework, natively handles discrete event-induced multi-modal dynamics. By directly modeling the intrinsic per-mode ODE vector fields, it flexibly accommodates variable-length and irregular trajectories.  This advances over fixed-step single-mode discrete networks (RNN/LSTM) that learn only a uniform step map conflating dynamics with sampling error.


The learning efficiency is then analyzed. The average training curves from 10 independent trials for all six configurations are shown in Fig. \ref{fig: learning efficiency}(e), where models with the EA framework demonstrate faster convergence and lower final training error. Notably, PENODE reached an equivalent loss level about 40 epochs earlier than other methods. As shown in Fig. \ref{fig: learning efficiency}(f), the EA framework also significantly reduces model complexity by decreasing the required neuron numbers for all networks, especially with the most substantial reduction of over 75\% for NODE, revealing excellent compatibility between the EA framework and continuous-time models. 

\subsection{Sim2Real Transfer Capability of PENODE }

The section evaluates the PENODE capability in Sim2Real transfer tasks. A two-stage dc-dc converter is used as the case study, as shown in Fig. \ref{fig: platform}. Three scenarios are evaluated: 1) white-box where the full converter model is known, 2) gray-box where only the DAB model is known, and 3) black-box  where no physical model is available.

In each scenario, PENODE is compared with two prominent physics-informed methods: PINN that incorporates physical laws as a loss constraint and uses a unified approach, and PRNN that uses an LSTM to learn the residual of a discretized physical model and employs a two-stage process. The total number of neurons for all models is capped at 64, and they are trained on datasets of varying sizes to evaluate the impact of data availability on accuracy.

The test results with 25\% data are shown in Fig. \ref{fig: test results}. In the gray-box case, the physics term of PRNN deviates from the true solution because the two-stage training of PRNN is prone to finding a sub-optimal physical solution, which makes its second-stage RNN to learn a larger residual with errors that the physical model could have otherwise corrected. It not only increases the learning burden on RNN but also risks introducing spurious correlations, leading to inaccurate physical parameters and an impure data-driven model. In the black-box scenario, PRNN performed poorly without a physical model foundation. In contrast, PENODE demonstrated superior performance in both gray-box and black-box scenarios, especially in the gray-box out-of-domain tests, due to its strength in unified representation and training process for both the physics-based and data-driven components. Through unified learning, PENODE allows gradients to flow to both the physics and data terms simultaneously to find a Pareto-optimal solution and directs the network to learn a very "small" and "smooth" correction vector field, while the gradient only naturally flows to the NNs when a dynamic pattern cannot be captured by any combination of physical parameters. Thus, PENODE makes the learning process more robust and the results more physically sound, as shwon in Fig. \ref{fig: test results}.x.3 and \ref{fig: test results}.x.4. 
The in-domain and out-of-domain relative errors under varying data quantities are shown in Fig. \ref{fig: test results}(d) and \ref{fig: test results}(e), indicating that PENODE maintains the lowest error across all scenarios, especially in the gray-box and black-box cases. These findings demonstrate that PENODE has excellent adaptability and generalization capabilities for Sim2Real transfer. A summary comparison of all methods is provided in Table IV.


\subsection{Real-time Performance of PENODE-based EDT}

\begin{figure}[t] 
    \centering
    \vspace{-15pt} 
    \includegraphics[width=0.5\textwidth]{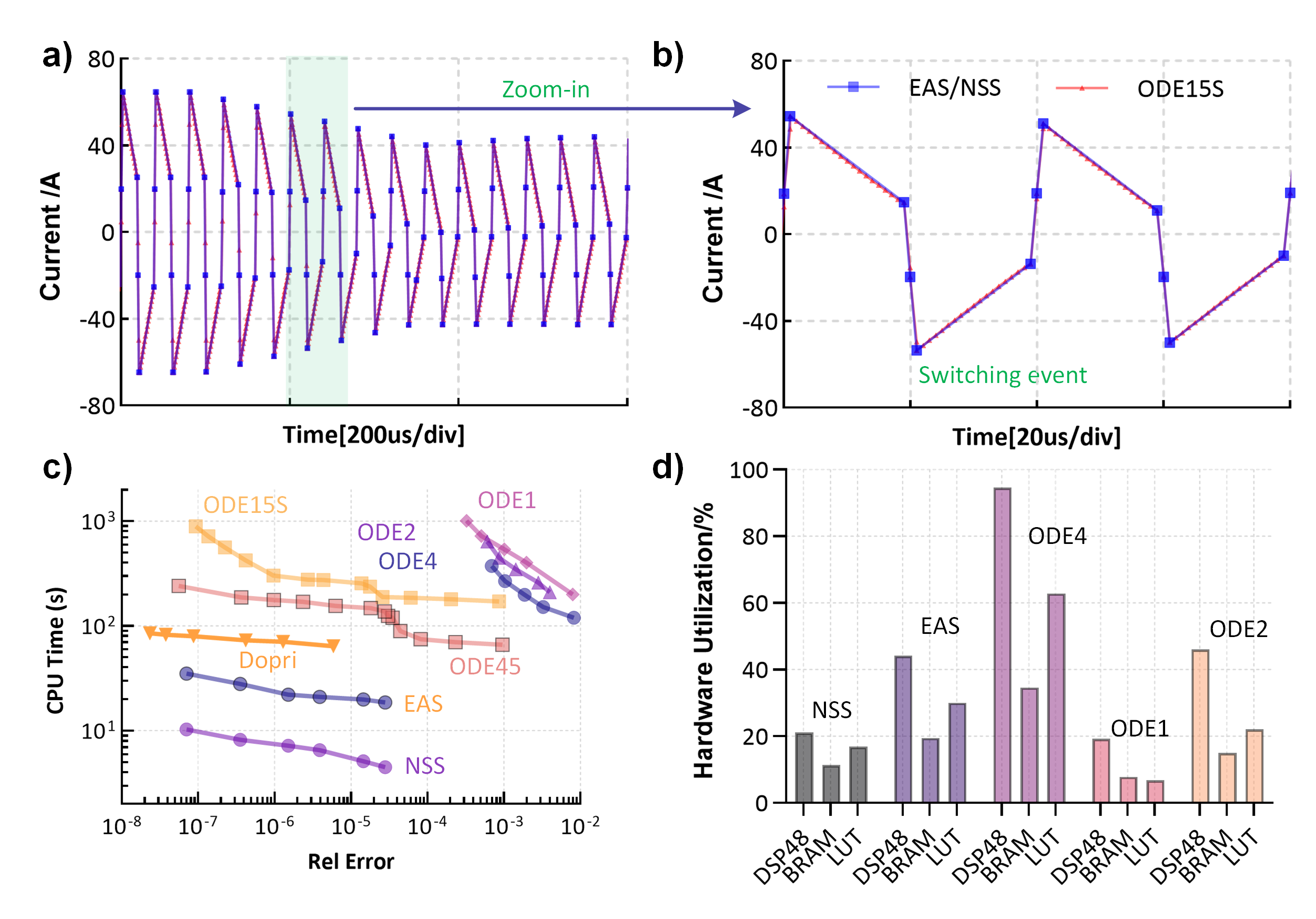} 
    \vspace{-25pt} 
    \caption{Evaluation of Real-time Performance for PENODE Model.(a). Time-domain simulation results of the trained PENODE model using different solvers; (b). zoom-in view of switching event region to compare EAS/NSS with ODE15S; 
    (c). Computation time versus relative error for various solvers, showing trade-offs between accuracy and efficiency; 
    (d). Hardware resource utilization of different solvers on FPGA, including DSP, BRAM, and LUT consumption.}
    \vspace{-12pt} 
    \label{fig: real_time_performance}
\end{figure}

In this section, the real-time performance of the PENODE method is evaluated from the perspective of its implementation as an EDT. 
As shown in Fig. \ref{fig: real_time_performance}, PENODE using the proposed event-driven solver is compared with the standard ode15s solver. During inference, the event-driven solver computes the state only at switching instants, which reduces the computational load and improves efficiency \cite{10236507, zheng2025neuralsubstitutesolverefficient}. Fig. \ref{fig: real_time_performance}(c) shows the computation time of PENODE with different solvers and error tolerances, indicating the flexibility of PENODE to allow it with efficient event-driven solvers when deployed on edge devices, thus achieving real-time performance while maintaining accuracy. Furthermore, the event-driven solver significantly reduces resource consumption, making it well-suited for resource-constrained edge environments. In contrast, models like PRNN are rigidly dependent on a fixed discrete step and require a full retraining when the step size is changed. 

\begin{figure}[t] 
    \centering
    \vspace{-15pt} 
    \includegraphics[width=0.5\textwidth]{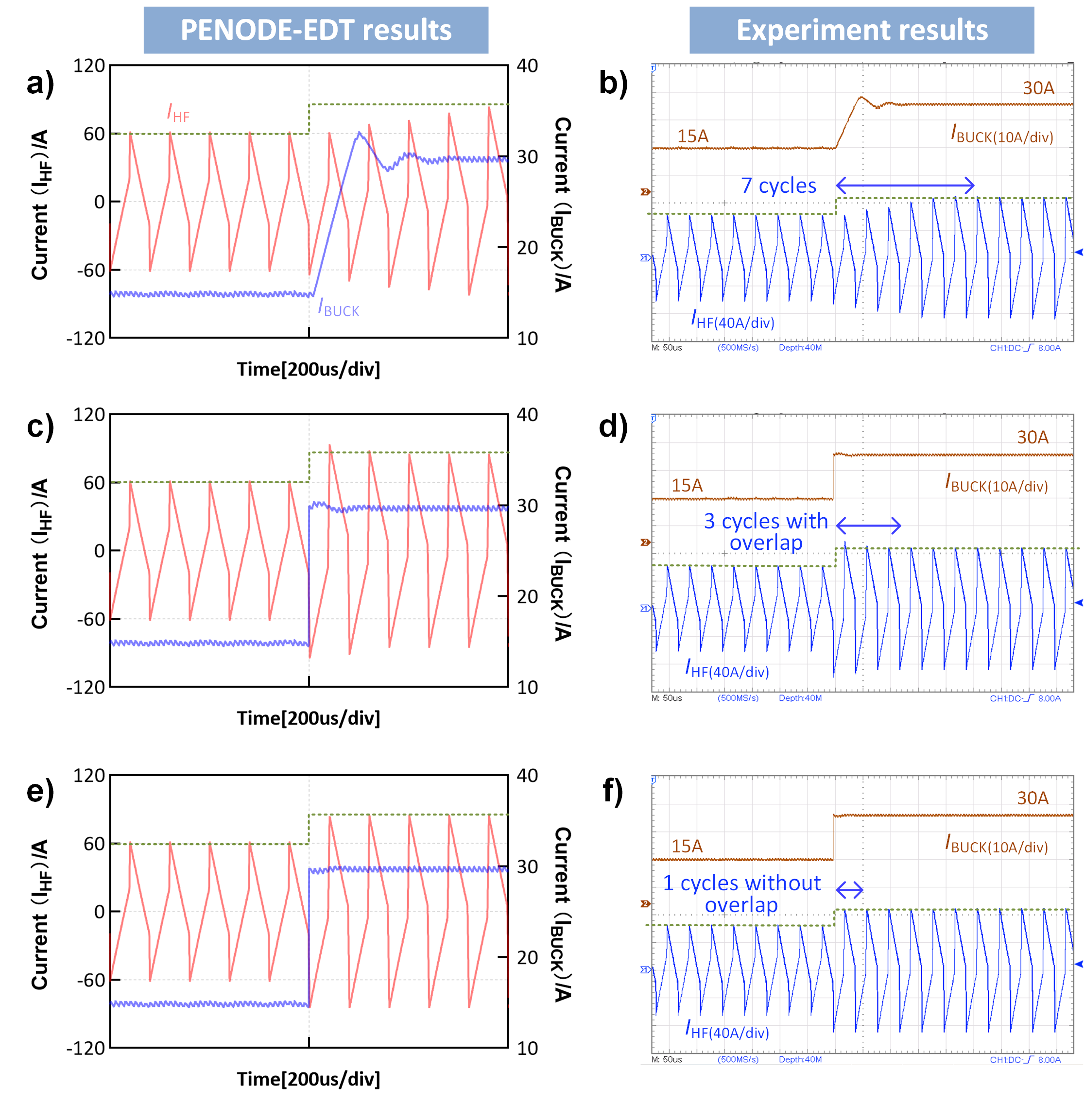} 
    \vspace{-25pt} 
    \caption{Real-time validation and control enhancement of PENODE-EDT on physical hardware. (a), (b) Real-time prediction results from PENODE-EDT under PI control. (c), (d) Sim2Sim EDT and experiment under MPC. (e), (f) Sim2Real EDT and experiment under MPC. }
    \vspace{-10pt} 
    \label{fig: real_time_performance_converter}
\end{figure}

An experiment on the physical converter has also been conducted. First, the EDT using the PENODE model processes the PWM signals in real time under a traditional PI controller, as shown in Fig. \ref{fig: real_time_performance_converter}(a) and (b). The EDT output closely matches the experimental measurements, confirming that the PENODE model is precise and can achieve switch-level dynamic tracking. Next, PENODE is implemented upon an advanced model predictive control (MPC) strategy that has faster dynamic responses than PI control but requires the sampling of the midpoint value of the high-frequency current in DAB from the EDT \cite{10043796, 11077776}. A comparative test has been conducted between an EDT based on the initial physical model (i.e., a "Sim2Sim" EDT) and an EDT based on the data-calibrated PENODE model (i.e., a "Sim2Real" EDT), as shown in Fig. \ref{fig: real_time_performance_converter}(c) through (f), revealing that the Sim2Real PENODE EDT achieves superior dynamic tracking performance than the Sim2Sim EDT that produces significant overshoot during dynamic responses due to model parameter mismatch. The experiment demonstrates that PENODE can not only perform real-time high-fidelity inference but also tangibly enhance the control performance of a physical system.

\section{Conclusion}\label{SectionVI}

This paper proposed a Physics-Embedded Neural ODEs method to achieve Sim-to-Real EDTs for PES. 
The PENODE can learn the multi-modal structure and the hybrid dynamics of PES directly from data through the event automata and model each identified operating mode using a continuous-time neural network to parameterize the governing ODEs directly. The unified design structurally embeds physical laws into a single model that supports end-to-end training. Furthermore, a complete cloud-to-edge workflow is established to ensure the model can be efficiently deployed onto resource-constrained edge devices. Experimental results have shown that PENODE significantly outperforms state-of-the-art benchmarks in accuracy across various scenarios and largely reduces the model complexity. This work validates a new paradigm for high-fidelity digital twins, which is efficient, accurate, physically interpretable, and flexibly deployable on edge hardware, paving the way for advanced real-time control applications.

\bibliographystyle{Bibliography/IEEEtranTIE}
\bibliography{Bibliography/IEEEabrv,Bibliography/BIB_xx-TIE-xxxx}

\begin{thebibliography}{10}
\providecommand{\url}[1]{#1}
\csname url@samestyle\endcsname
\providecommand{\newblock}{\relax}
\providecommand{\bibinfo}[2]{#2}
\providecommand{\BIBentrySTDinterwordspacing}{\spaceskip=0pt\relax}
\providecommand{\BIBentryALTinterwordstretchfactor}{4}
\providecommand{\BIBentryALTinterwordspacing}{\spaceskip=\fontdimen2\font plus
\BIBentryALTinterwordstretchfactor\fontdimen3\font minus \fontdimen4\font\relax}
\providecommand{\BIBforeignlanguage}[2]{{%
\expandafter\ifx\csname l@#1\endcsname\relax
\typeout{** WARNING: IEEEtran.bst: No hyphenation pattern has been}%
\typeout{** loaded for the language `#1'. Using the pattern for}%
\typeout{** the default language instead.}%
\else
\language=\csname l@#1\endcsname
\fi
#2}}
\providecommand{\BIBdecl}{\relax}
\BIBdecl

\bibitem{10070105}
F.~Blaabjerg \emph{et~al.}, ``Power electronics technology for large-scale renewable energy generation,'' \emph{Proc. IEEE}, vol. 111, no.~4, pp. 335--355, 2023.

\bibitem{9535421}
Z.~Tang \emph{et~al.}, ``Power electronics: The enabling technology for renewable energy integration,'' \emph{CSEE Journal of Power and Energy Systems}, vol.~8, no.~1, pp. 39--52, 2022.

\bibitem{9784425}
F.~Arraño-Vargas \emph{et~al.}, ``Modular design and real-time simulators toward power system digital twins implementation,'' \emph{IEEE Transactions on Industrial Informatics}, vol.~19, no.~1, pp. 52--61, 2023.

\bibitem{9950705}
H.~Chen \emph{et~al.}, ``Digital twin techniques for power electronics-based energy conversion systems: A survey of concepts, application scenarios, future challenges, and trends,'' \emph{IEEE Industrial Electronics Magazine}, vol.~17, no.~2, pp. 20--36, 2023.

\bibitem{9921407}
D.~Gebbran \emph{et~al.}, ``Cloud and edge computing for smart management of power electronic converter fleets: A key connective fabric to enable the green transition,'' \emph{IEEE Industrial Electronics Magazine}, vol.~17, no.~2, pp. 6--19, 2023.

\bibitem{11077776}
J.~Zheng \emph{et~al.}, ``Cognitive digital twins-based model predictive control for high-frequency power converters,'' \emph{IEEE Transactions on Industrial Electronics}, pp. 1--12, 2025.

\bibitem{9525187}
S.~Shao \emph{et~al.}, ``Modeling and advanced control of dual-active-bridge dc–dc converters: A review,'' \emph{IEEE Transactions on Power Electronics}, vol.~37, no.~2, pp. 1524--1547, 2022.

\bibitem{plecsmanual}
\BIBentryALTinterwordspacing
{Plexim GmbH}, \emph{PLECS User Manual}, Plexim GmbH, 2025, [Accessed: March 3, 2025]. [Online]. Available: \url{https://www.plexim.com/files/plecsmanual.pdf}
\BIBentrySTDinterwordspacing

\bibitem{4303291}
A.~Davoudi \emph{et~al.}, ``Parasitics realization in state-space average-value modeling of pwm dc–dc converters using an equal area method,'' \emph{IEEE Transactions on Circuits and Systems I: Regular Papers}, vol.~54, no.~9, pp. 1960--1967, 2007.

\bibitem{10057400}
H.~Ge \emph{et~al.}, ``Modeling methodology based on fast and refined neural networks for non-isolated dc–dc converters with configurable parameter settings,'' \emph{IEEE Journal on Emerging and Selected Topics in Circuits and Systems}, vol.~13, no.~2, pp. 617--628, 2023.

\bibitem{7222462}
B.~X. Li \emph{et~al.}, ``Low sampling rate online parameters monitoring of dc–dc converters for predictive-maintenance using biogeography-based optimization,'' \emph{IEEE Transactions on Power Electronics}, vol.~31, no.~4, pp. 2870--2879, 2016.

\bibitem{10342664}
H.~Xu \emph{et~al.}, ``Numerical derivative-based flexible integration algorithm for power electronic systems simulation considering nonlinear components,'' \emph{IEEE Transactions on Industrial Electronics}, vol.~71, no.~9, pp. 10\,761--10\,771, 2024.

\bibitem{9819434}
J.~Zheng \emph{et~al.}, ``An event-driven real-time simulation for power electronics systems based on discrete hybrid time-step algorithm,'' \emph{IEEE Transactions on Industrial Electronics}, vol.~70, no.~5, pp. 4809--4819, 2023.

\bibitem{10236507}
Z.~Jialin \emph{et~al.}, ``Mpsoc-based dynamic adjustable time-stepping scheme with switch event oversampling technique for real-time hil simulation of power converters,'' \emph{IEEE Transactions on Transportation Electrification}, vol.~10, no.~2, pp. 3560--3575, 2024.

\bibitem{9832797}
J.~Zheng \emph{et~al.}, ``An event-driven parallel acceleration real-time simulation for power electronic systems without simulation distortion in circuit partitioning,'' \emph{IEEE Transactions on Power Electronics}, vol.~37, no.~12, pp. 15\,626--15\,640, 2022.

\bibitem{DBLP}
M.~Poli \emph{et~al.}, ``Neural hybrid automata: Learning dynamics with multiple modes and stochastic transitions,'' \emph{arxiv}, 2021.

\bibitem{7129353}
T.~Kamel \emph{et~al.}, ``Capacitor aging detection for the dc filters in the power electronic converters using anfis algorithm,'' in \emph{2015 IEEE 28th Canadian Conference on Electrical and Computer Engineering (CCECE)}, pp. 663--668, 2015.

\bibitem{9067098}
G.~Rojas-Dueñas \emph{et~al.}, ``Black-box modelling of a dc-dc buck converter based on a recurrent neural network,'' in \emph{2020 IEEE International Conference on Industrial Technology (ICIT)}, pp. 456--461, 2020.

\bibitem{9813409}
M.~Micev \emph{et~al.}, ``Artificial neural network-based nonlinear black-box modeling of synchronous generators,'' \emph{IEEE Transactions on Industrial Informatics}, vol.~19, no.~3, pp. 2826--2837, 2023.

\bibitem{9255375}
P.~Qashqai \emph{et~al.}, ``Modeling power electronic converters using a method based on long-short term memory (lstm) networks,'' in \emph{IECON 2020 The 46th Annual Conference of the IEEE Industrial Electronics Society}, pp. 4697--4702, 2020.

\bibitem{10463542}
X.~Li \emph{et~al.}, ``Temporal modeling for power converters with physics-in-architecture recurrent neural network,'' \emph{IEEE Transactions on Industrial Electronics}, vol.~71, no.~11, pp. 14\,111--14\,123, 2024.

\bibitem{NODE}
T.~Q. Chen \emph{et~al.}, ``Neural ordinary differential equations,'' \emph{NIPS}, vol. abs/1806.07366, 2018.

\bibitem{9779551}
S.~Zhao \emph{et~al.}, ``Parameter estimation of power electronic converters with physics-informed machine learning,'' \emph{IEEE Transactions on Power Electronics}, vol.~37, no.~10, pp. 11\,567--11\,578, 2022.

\bibitem{9693757}
M.~Lebedev \emph{et~al.}, ``A survey of open-source tools for fpga-based inference of artificial neural networks,'' in \emph{2021 Ivannikov Memorial Workshop (IVMEM)}, pp. 50--56, 2021.

\bibitem{DBLP:journals/corr/abs-2004-06632}
\BIBentryALTinterwordspacing
L.~Datta, ``A survey on activation functions and their relation with xavier and he normal initialization,'' \emph{CoRR}, vol. abs/2004.06632, 2020. [Online]. Available: \url{https://arxiv.org/abs/2004.06632}
\BIBentrySTDinterwordspacing

\bibitem{zheng2025neuralsubstitutesolverefficient}
\BIBentryALTinterwordspacing
J.~Zheng \emph{et~al.}, ``Neural substitute solver for efficient edge inference of power electronic hybrid dynamics,'' 2025. [Online]. Available: \url{https://arxiv.org/abs/2507.03144}
\BIBentrySTDinterwordspacing

\bibitem{10043796}
E.~Zafra \emph{et~al.}, ``Computationally efficient sphere decoding algorithm based on artificial neural networks for long-horizon fcs-mpc,'' \emph{IEEE Transactions on Industrial Electronics}, vol.~71, no.~1, pp. 39--48, 2024.

\end{thebibliography}

\vspace{-60pt}
\begin{IEEEbiography}[{\includegraphics[width=1in,height=1.25in,clip,keepaspectratio]{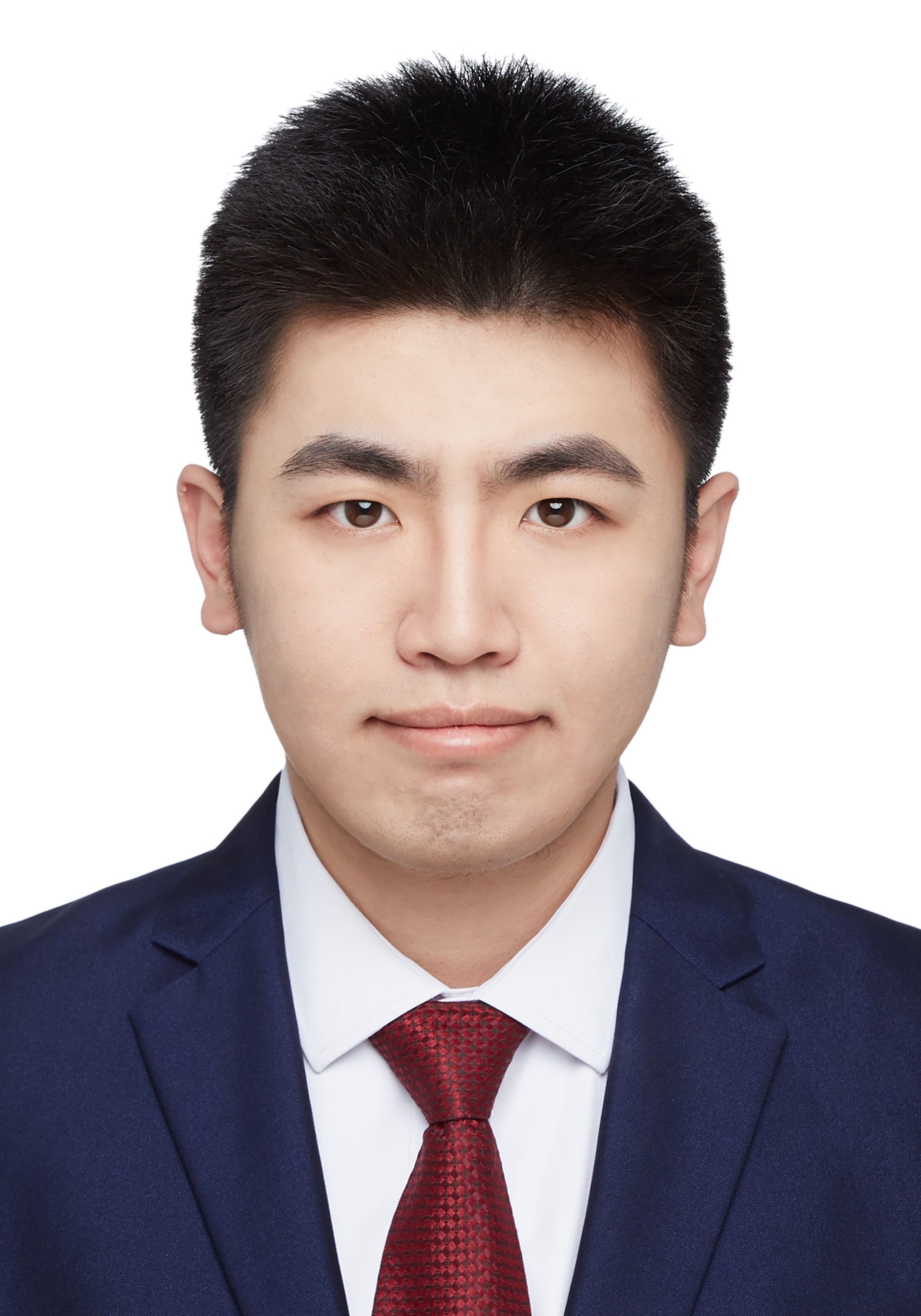}}]
{Jialin Zheng} (Member, IEEE) received the B.S. degree in electrical engineering in 2019 from Beijing Jiaotong University, Beijing, China, and the Ph.D. degree in electrical engineering in 2024 from Tsinghua University, Beijing, China. From 2024,  He start to be Research Fellow with Purdue University, the USA. His research interests include real-time simulation, machine learning, and digital twins technology in power electronics.
\end{IEEEbiography}

\vspace{-40pt}
\begin{IEEEbiography}[{\includegraphics[width=1in,height=1.25in,clip,keepaspectratio]{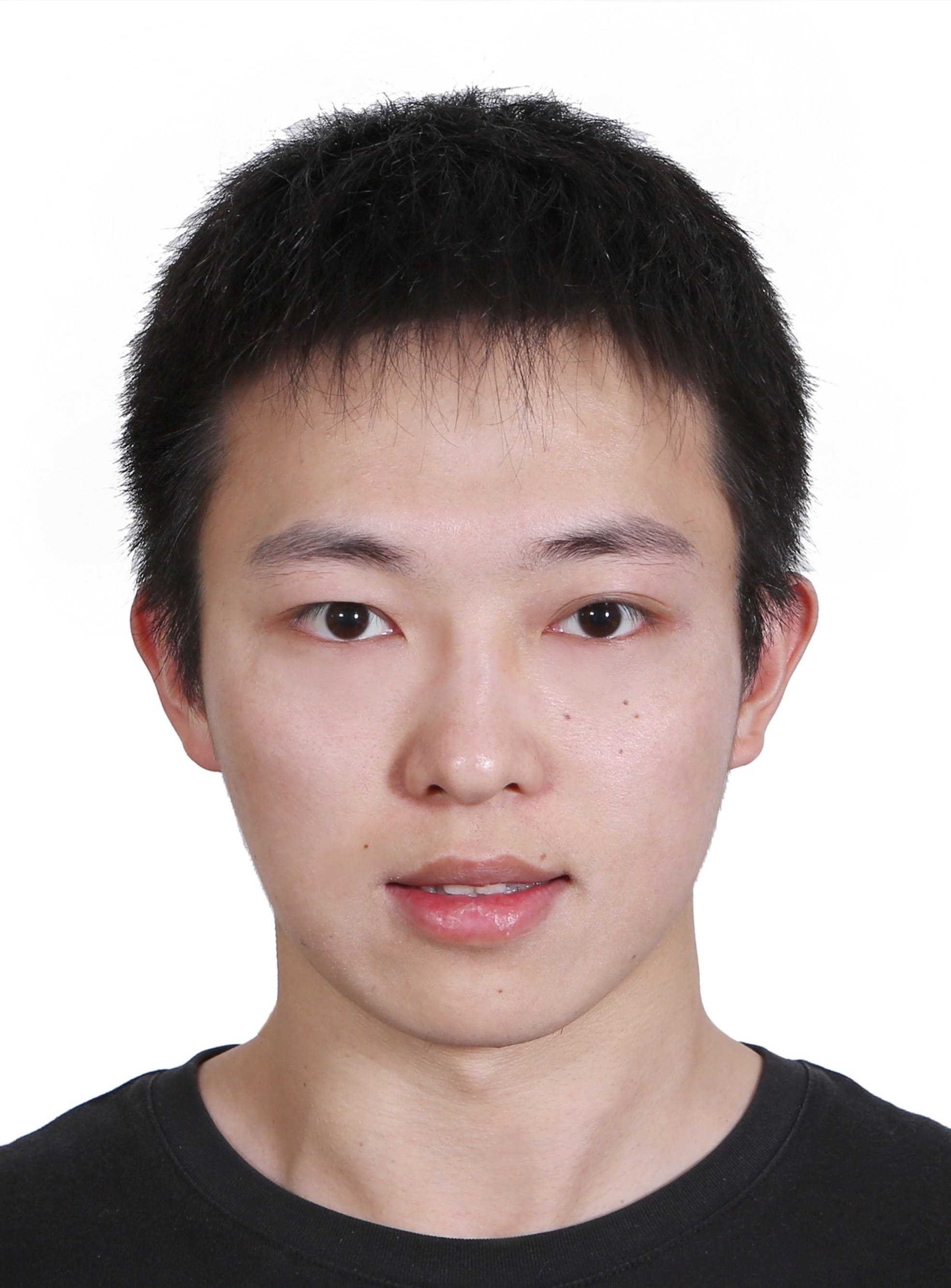}}]
{Haoyu Wang}(Student member, IEEE) received the B.S. degree in electrical engineering and automation from Shanghai Jiao Tong University, Shanghai, China, in 2020, and the M.S. degree in electrical engineering from Tsinghua University, Beijing, China, in 2023. He is currently working toward the Ph.D. degree in electrical engineering at UT Austin, TX. 
His current research interests include dc-dc converters, and AI-based design and control of power converters.
\end{IEEEbiography}

\vspace{-40pt}
\begin{IEEEbiography}[{\includegraphics[width=1in,height=1.25in,clip,keepaspectratio]{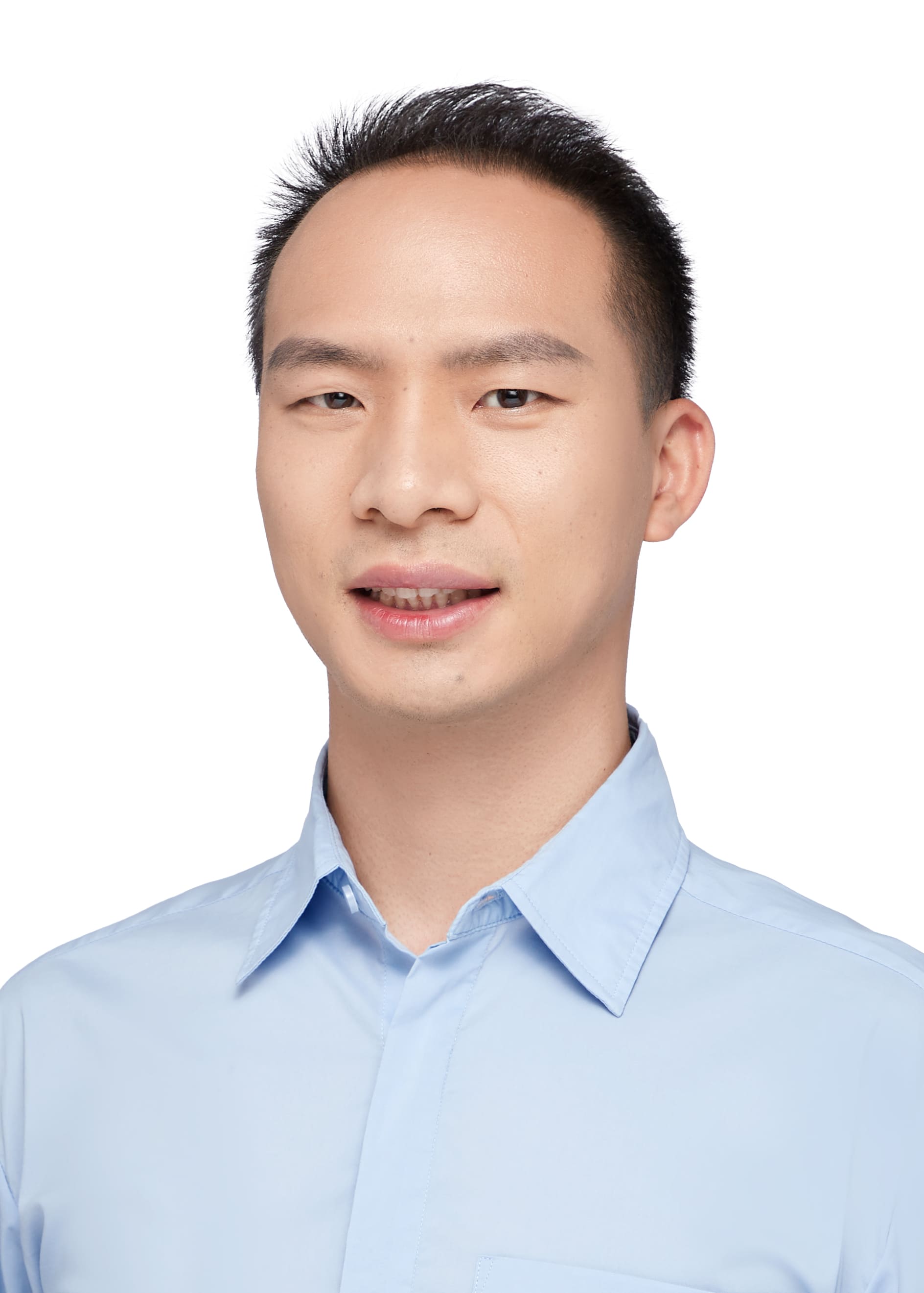}}]
{Yangbin Zeng} (Member, IEEE)  received the B.Sc. degree in building electrical and intelligent from Xiangtan University in 2015, and the Ph.D. degree in electrical engineering from Beijing Jiaotong University in 2021. From 2021 to 2024, he was a postdoc at Tsinghua University. He is currently an Associate Professor with South China University of Technology, Guangzhou, China. His current research interests include power electronics control.
\end{IEEEbiography}

\vspace{-40pt}
\begin{IEEEbiography}[{\includegraphics[width=1in,height=1.25in,clip,keepaspectratio]{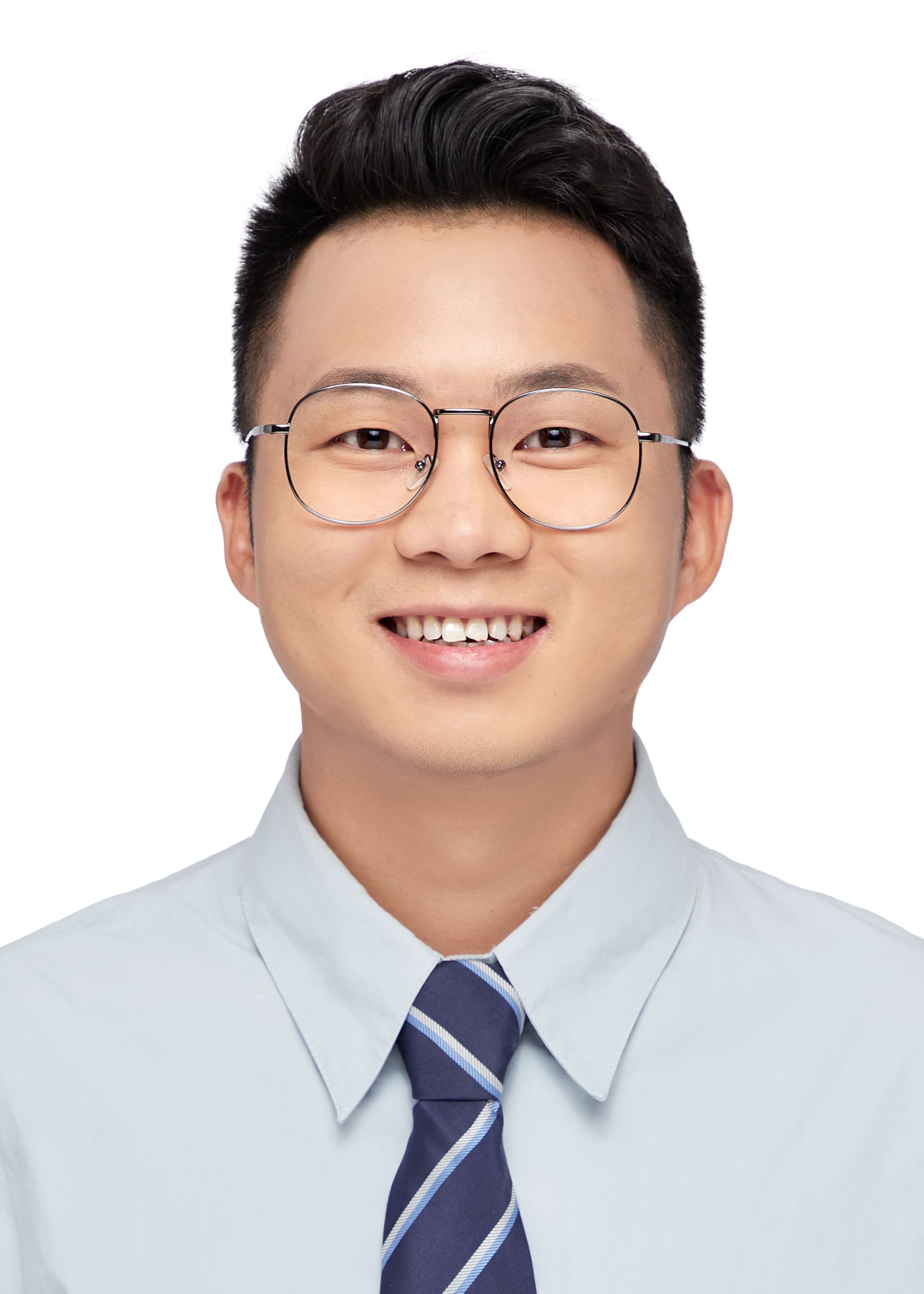}}]
{Di Mou} (Member, IEEE) was born in Lichuan, Hubei Province, China, in 1994. He received the B.S. degree from the Three Gorge University, Yichang, China, in 2017, and the Ph.D. degree from the Chongqing University, Chongqing, China, in 2021, both in electrical engineering. He is currently a Postdoctoral Fellow with the Tsinghua University, Beijing, China. His research interests include multiport power electronic transformers.
\end{IEEEbiography}

\vspace{-40pt}
\begin{IEEEbiography}[{\includegraphics[width=1in,height=1.25in,clip,keepaspectratio]{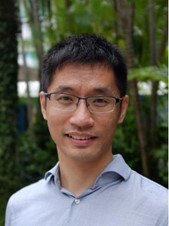}}]
{Xin Zhang} (Senior Member, IEEE) received the Ph.D. degree in automatic control and systems engineering from the University of Sheffield, Sheffield, U.K., in 2016. From 2017 to 2020, he was an Assistant Professor with Nanyang Technological University. After that, he is currently a Professor with Zhejiang University. His research interests include power electronics and advanced control theory.  He is the Associated Editor of IEEE TIE, JESTPE, etc. 
\end{IEEEbiography}

\vspace{-40pt}
\begin{IEEEbiography}[{\includegraphics[width=1in,height=1.25in,clip,keepaspectratio]{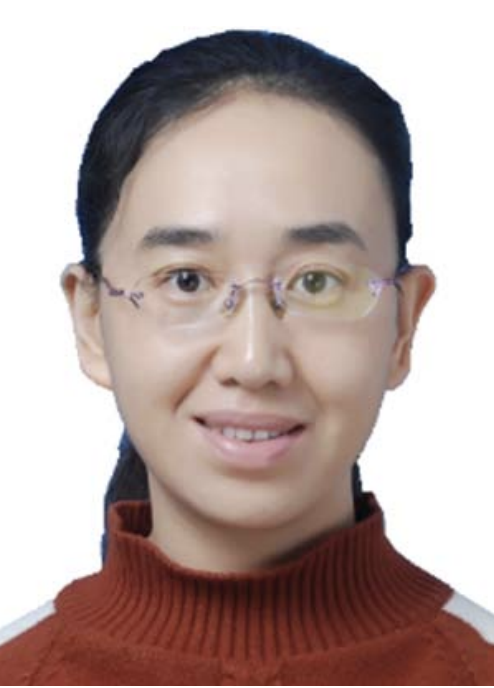}}]
{Hong Li} (Senior Member, IEEE) received the
B.E. degree from Taiyuan University of Technology, Taiyuan, China, in 2002, the M.E. degree from the South China University of Technology, Guangzhou, China, in 2005, and the Ph.D. degree from FernUniversitat in Hagen, Hagen, Germany, in 2009, all in electrical engineering. Her research interests include nonlinear modeling in power electronics. She is the Associated Editor of IEEE TIE, TPEL, OJIES, etc. 
\end{IEEEbiography}

\vspace{-40pt}
\begin{IEEEbiography}[{\includegraphics[width=1in,height=1.25in,clip,keepaspectratio]{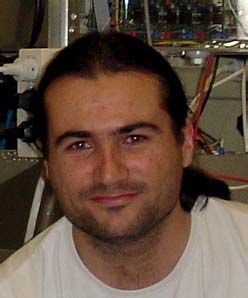}}]
{Sergio Vazquez} (Fellow, IEEE) was born in Seville, Spain, in 1974. He received the M.S. and Ph.D. degrees in industrial engineering from the University of Seville (US), Seville, Spain, in 2006 and 2010, respectively. His research interests include power electronics systems modeling. He is involved in the Energy Storage Technical Committee of the IEEE Industrial Electronics Society and is currently an Co-EIC of the IEEE TIE.
\end{IEEEbiography}

\vspace{-40pt}
\begin{IEEEbiography}[{\includegraphics[width=1in,height=1.25in,clip,keepaspectratio]{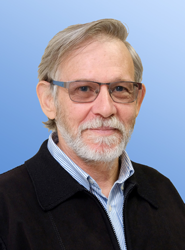}}]
{Leopoldo G. Franquelo} (Life Fellow, IEEE) was born in Malaga, Spain. He received the M.Sc. and Ph.D. degrees in electrical engineering from the Universidad de Sevilla, Seville, Spain, in 1977 and 1980, respectively. His research interests include power electronics systems modeling.  In 2012 and 2015, he was the recipient of the Eugene Mittelmann Outstanding Research Achievement Award and the Anthony J. Hornfeck Service Award from IEEE-IES, respectively. In IEEE Transactions on Industrial Electronics, he became an Associate Editor in 2007, Co-Editor-in-Chief in 2014, and the Editor-in-Chief in 2016. He was the President of the IES (2010–2011) and is an IES AdCom Life Member.
\end{IEEEbiography}

\end{document}